\documentclass[journal]{IEEEtran}
\usepackage{amsmath,amsfonts}
\usepackage[ruled,vlined]{algorithm2e}  
\usepackage{array}
\usepackage{textcomp}
\usepackage{stfloats}
\usepackage{url}
\usepackage{verbatim}
\usepackage{graphicx}
\def\BibTeX{{\rm B\kern-.05em{\sc i\kern-.025em b}\kern-.08em
    T\kern-.1667em\lower.7ex\hbox{E}\kern-.125emX}}
\usepackage{balance}

\usepackage{subcaption}
\usepackage{multirow}
\usepackage{booktabs}
\usepackage{xcolor}
\usepackage{tcolorbox} 

\SetArgSty{textnormal}

\begin{document}

\title{Local-Minimum Escaper: Programmatic Subgoal Generation for Robust Navigation in Unknown Environments}
\author{Yin Gu, Xinming Zhang, Shanze Wang, Siwei Cheng, Wei Zhang
\thanks{Yin Gu is with Eastern Institute of Technology, Ningbo, China, and also with Tsinghua University, China. E-mail: gy128@mail.ustc.edu.cn.}

\thanks{Xinming Zhang, Siwei Cheng and Wei Zhang (corresponding author) are with Eastern Institute of Technology, Ningbo, China. E-mail: \{xm\_zhang, chengsiwei\}@mail.ustc.edu.cn, zhw@eitech.edu.cn.}

\thanks{Shanze Wang is with Eastern Institute of Technology, Ningbo, China, and also with the Department of Aeronautical and Aviation Engineering, Polytechnic University, Hong Kong. E-mail: shanze.wang@connect.polyu.hk.}
}

\markboth{Journal of \LaTeX\ Class Files,~Vol.~18, No.~9, September~2020}
{How to Use the IEEEtran \LaTeX \ Templates}

\maketitle

\begin{abstract}

Mapless navigation in unknown and partially observable environments remains challenging for mobile robots, particularly when local minima prevent the robot from making progress toward its goal. Existing local navigation methods often lack an explicit mechanism for escaping such situations, while deep reinforcement learning (DRL) approaches typically learn recovery behaviors implicitly through reward design and policy optimization. In this work, we propose \textbf{LME} (Local-Minimum Escaper), a programmatic hierarchical framework that explicitly generates and reasons subgoals to guide robots out of local-minimum regions. LME operates solely on local observations and selects candidate subgoals using interpretable heuristic criteria that account for both surrounding obstacle geometry and candidate-location safety. A local planner then generates low-level motion commands toward the selected subgoal. This design enables LME to handle environments both with and without local minima within a unified framework, while remaining independent of the underlying local planner and requiring no additional training.
Extensive experiments in simulated and real-world environments demonstrate that LME provides robust navigation performance and generalizes to challenging unseen scenarios. Furthermore, the generated subgoals can be used to guide different local planners, substantially improving their ability to escape local minima. Successful deployments on both differential-drive and quadruped robots further demonstrate the practical applicability and generality of the proposed framework.

\end{abstract}


\section{Introduction}

\IEEEPARstart{N}avigation in unknown environments is fundamental to many transportation applications, particularly in dynamic and unstructured scenarios such as search and rescue~\cite{calisi2005autonomous}, autonomous exploration~\cite{sodiya2024ai}, and service robotics~\cite{yu2024soft}. In such settings, a robot must continuously make motion decisions based on incomplete and potentially changing observations while safely reaching a designated goal~\cite{jian2026crar,wang2025navbest}. Conventional navigation and path-planning approaches typically rely on an explicit representation of the environment or predefined motion heuristics. For example, graph- and sampling-based methods such as A*~\cite{likhachev2005anytime} and RRT~\cite{islam2012rrt} compute collision-free paths based on a known or constructed map. In contrast, local and reactive methods directly generate motion commands from the robot's current state and local observations. The Dynamic Window Approach (DWA)~\cite{fox_dynamic_1997,dobrevski2024dynamic} selects control commands from a feasible velocity set that satisfies kinematic constraints while avoiding collisions and making progress toward the goal. Artificial Potential Fields (APF)~\cite{khatib1986real} guide the robot by modeling the goal as an attractive force and obstacles as repulsive forces. While these methods have demonstrated strong performance in structured environments, their reliance on explicit maps, or manually tuned parameters can limit their adaptability to complex and previously unseen environments.

In recent years, Deep Reinforcement Learning (DRL)~\cite{zhang2022ipaprec,Li2022,zhang2025drl,hu2024deep,xie_drl-vo_2023} has emerged as a promising data-driven paradigm for mapless navigation. By continuously interacting with the environment, a DRL agent can learn a navigation policy that maps sensor observations and the robot's state directly to motion commands, optimizing for collision-free and goal-directed behavior without requiring an explicit global map. Building upon this capability, recent studies have further investigated navigation in challenging scenarios involving dynamic obstacles~\cite{van2011reciprocal,xie_drl-vo_2023}, narrow passages~\cite{xiao2023autonomous,wang2025maer}, and other complex environmental constraints. These advances demonstrate the potential of DRL to provide adaptive navigation behaviors directly from local observations.

However, relatively little attention has been paid to local minima in navigation under unknown environments. Without a global map, a robot can easily become trapped in a locally optimal state when simultaneously approaching the goal and avoiding obstacles. Without an effective escape mechanism, the robot may remain indefinitely stuck in the same region.
Recent DRL-based methods have attempted to address this problem by learning escape behaviors through carefully designed rewards and training procedures. For example, HIT-RL~\cite{jang2021hindsight} heuristically generates intermediate targets and employs hindsight experience replay to escape local-minimum regions and dead ends. T-DRL~\cite{zhu2025t} generates local goals and rewards the robot for moving toward them, thereby alleviating local minima caused by dynamic pedestrians. Other approaches~\cite{hu2024deep} encourage exploration by penalizing previously visited regions and rewarding diverse local exploration. While effective, these methods address local minima indirectly through policy learning and reward engineering. This typically requires balancing multiple reward terms and may limit generalization beyond the training environments. Moreover, the resulting policies are often difficult to interpret and verify~\cite{rudin2019stop}.

In this work, we propose \textbf{LME} (Local-Minimum Escaper), a programmatic hierarchical framework that explicitly identifies promising subgoals for escaping local minima. Rather than implicitly learning an escape policy, LME explicitly reasons about where the robot should move to escape a locally trapped region. The high-level program generates subgoals from local observations based on a set of interpretable heuristic criteria, while a local planner computes the low-level actions required to reach the selected subgoal. These criteria jointly consider the geometric structure of surrounding obstacles and the safety of candidate locations, enabling the robot to identify subgoals that guide it around obstacles and out of local minima.
LME has three key advantages. First, it relies solely on local observations and does not require a global map. Second, its explicit subgoal generation enables the robot to handle both environments with and without local minima within a unified framework. Third, its programmatic structure makes the decision-making process interpretable and easy to analyze. Importantly, LME is also local-planner-agnostic: it can be seamlessly integrated with existing local planners by replacing their final goal with the generated subgoals, without requiring any retraining of the underlying planner.
Extensive experiments in simulated and real-world environments demonstrate that LME consistently improves navigation performance and generalization, outperforming both conventional navigation methods and recent state-of-the-art DRL-based approaches.

\section{Related Work}

\subsection{Local and Reactive Navigation}

Local and reactive navigation methods generate motion commands directly from the robot's current state and local observations, and therefore provide an important foundation for navigation in partially known or unknown environments. Representative approaches include the Dynamic Window Approach (DWA)~\cite{fox_dynamic_1997}, Elastic Band (E-Band)~\cite{quinlanElasticBandsConnecting1993}, Vector Field Histogram (VFH)~\cite{borenstein1991vector,ulrich1998vfh+}, and Artificial Potential Fields (APF)~\cite{khatib1986real}. DWA selects feasible velocity commands within a dynamically admissible velocity space, while E-Band locally deforms a reference path to maintain obstacle clearance and smoothness. VFH transforms local range observations into a polar obstacle-density representation and selects a collision-free motion direction. APF, in contrast, formulates navigation as the interaction between attractive and repulsive potential fields.
Classical reactive strategies also include the family of Bug algorithms~\cite{lumelsky1986dynamic,lumelsky1987path}, which navigate using local sensing and obstacle-boundary following without a global map.


\subsection{DRL for Mapless Navigation}
Deep Reinforcement Learning (DRL) has emerged as a powerful framework for mapless navigation, learning policies that directly map local sensory observations and robot state to motion commands without requiring an explicit global map. Early work demonstrated that DRL policies could learn collision-free goal-directed behaviors from laser observations and relative goal information~\cite{Pfeiffer2018,zhelo2018curiosity,zhang2022ipaprec,marchesini2020discrete}. 
Subsequent studies have investigated richer sensory representations, recurrent architectures, improved reward functions, and training strategies to improve navigation performance and generalization~\cite{Li2022,xie_drl-vo_2023,zhang2025drl,cai2024deep,wang2025maer,Wang2023,li2023context}.

Nevertheless, local minima remain a challenging case for mapless DRL navigation. In a locally trapped configuration, the action that immediately improves the distance to the final goal may not lead to a globally feasible route. The robot may therefore need to temporarily move away from the goal and explore alternative directions. This creates an inherent tension between goal-directed reward design and the exploratory behavior required for escaping local minima.

\subsection{Local-Minimum-Aware Mapless Navigation}

Several studies have explicitly investigated local-minimum or dead-end scenarios in mapless navigation. One representative approach is HIT-RL~\cite{jang2021hindsight}, which introduces hindsight intermediate targets to encourage the robot to take detours toward the final destination using recent local observations. Rather than directly optimizing motion toward the final goal, the method temporarily guides the agent toward intermediate targets, providing a mechanism for escaping locally trapped configurations. 
More recent methods have incorporated exploration mechanisms into DRL to alleviate local minima. For example, Hu et al.~\cite{hu2024deep} encourage the agent to explore previously less-visited regions by incorporating additional exploration-related rewards into SAC~\cite{haarnoja2018soft}. Similarly, Gao et al.~\cite{gao2024mapless} propose a hierarchical DRL framework in which the high-level policy is trained with extrinsic and intrinsic rewards based on novelty and memory, while an LSTM is employed to improve the agent's ability to reason about previously visited regions. Such approaches demonstrate that exploration and memory can substantially improve the ability of learned policies to recover from locally trapped configurations. More recently, hierarchical navigation frameworks have also investigated high-level subgoal generation for local-minimum avoidance, using environmental information to update subgoals while a low-level policy performs motion planning~\cite{gao2025hierarchical}.

Although these approaches demonstrate promising solutions to local-minimum navigation, their escape mechanisms are primarily learned through reward engineering, or additional exploration mechanisms. Consequently, the desired escape behavior is encoded implicitly in the learned policy, whose black-box nature makes the resulting decisions difficult to interpret and verify~\cite{glanois2021survey}.

In contrast to the DRL-Based approaches, we do not design complex reward functions or exploration mechanisms to indirectly incentivize escaping local optima. Instead, our method encodes local-optima avoidance directly at the policy level. Moreover, LME can be combined with different local planners without retraining the underlying navigation policy.

\section{LME Method}

\begin{figure}[t]
	\centering
	\includegraphics[width=0.7\columnwidth]{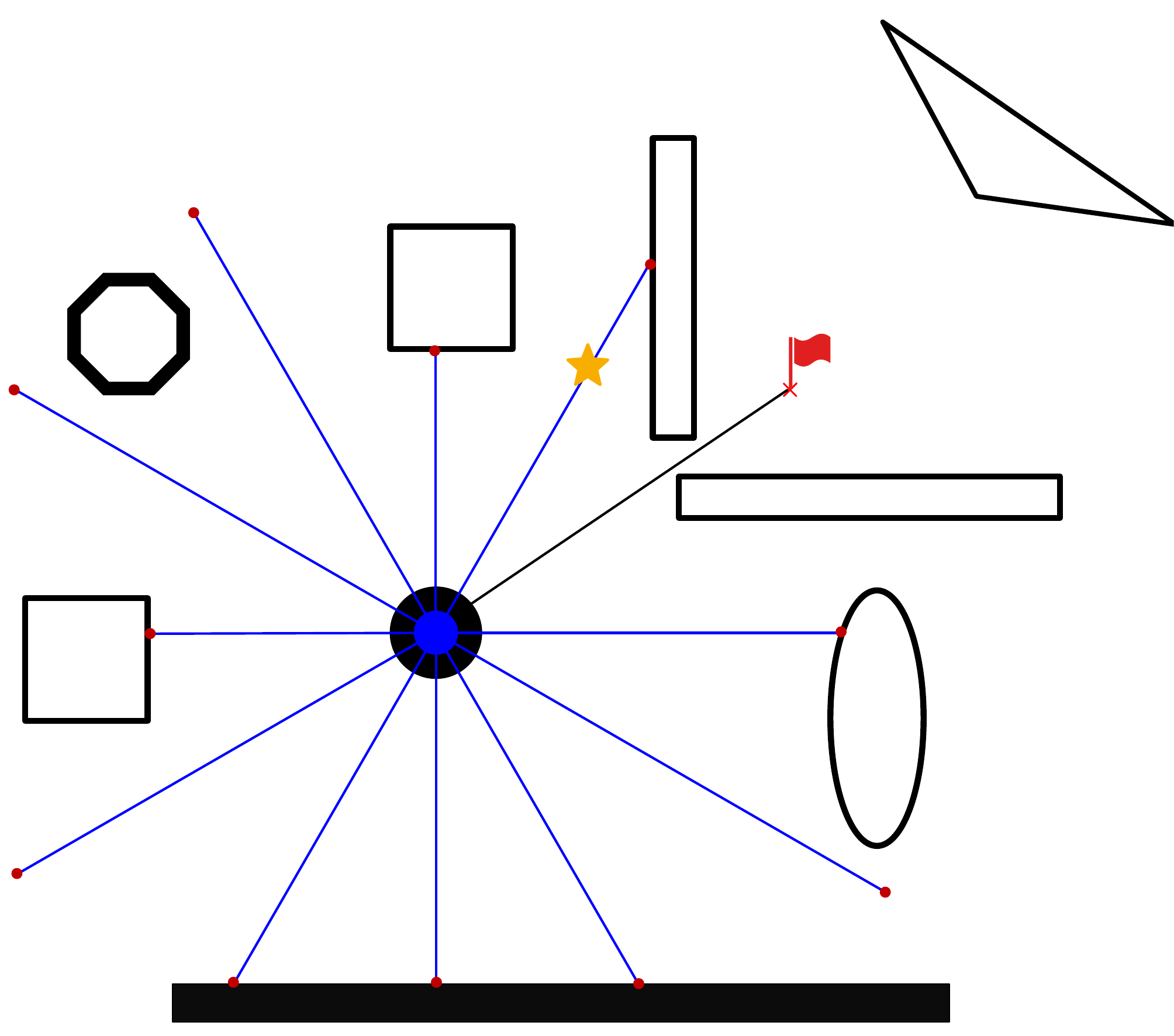}
	\caption{An illustration of the mapless navigation task. The blue lines represent the laser beams from the LiDAR. The red flag icon marks the final goal, while the several yellow stars represent the potential subgoals generated by LME.} 
	\label{example}
\end{figure}

\subsection{Problem Formulation}

We formulate mapless navigation as a sequential decision-making problem. Consider a robot equipped with a laser scanner navigating toward a designated target in an unknown environment without access to a global map, as illustrated in Fig.~\ref{example}. Following the terminology commonly used in reinforcement learning (RL), we refer to the decision-making program as a \textit{policy}.

At each time step, the policy’s observation is represented as
$\{\textit{scan}, p_T, v, \omega, v_\text{max}, \omega_\text{max}, \alpha_v, \alpha_\omega\}$,
where $\textit{scan}$ denotes the laser scan points, $p_T$ represents the relative target position, and $v$ and $\omega$ are the robot’s current linear and angular velocities, respectively. $(v_{\max},\omega_{\max})$ specify the maximum linear and angular velocity limits, while $(\alpha_v,\alpha_\omega)$ denote the corresponding maximum acceleration limits. Based on the current observation,
the policy outputs an action, which is a continuous set of linear and angular velocities. After executing, the next observation is updated with new sensory input.

\textbf{Reward.} LME adopts a sparse reward that depends on the final result of one episode:
\begin{equation}
R =
\begin{cases}
1 - \frac{L}{L_{\max}}, & \text{if success}, \\
R_c, & \text{if crash}, \\
R_t, & \text{if timeout},
\end{cases}
\end{equation}
where $L$ denotes the number of steps taken in the episode, and $L_{\max}$ is the predefined maximum number of steps per episode. The terms $R_c$ and $R_t$ are negative penalty values (e.g., $-0.2$), introduced to penalize undesirable outcomes such as collisions and timeouts. The objective of our method is to find an optimal policy that maximizes the expected $R$.

\subsection{Policy Framework}

\begin{algorithm}[t]
	\caption{Policy framework of LME}
	\KwIn{previous subgoal, current state}
	\KwOut{action}
	\tcp{\textbf{Subgoal Generation}}
	\tcp{Randomly sample N points}
    $H = [p_1, p_2, \dots, p_N]$ \\
    $[S_1, S_2, \dots, S_N] = \texttt{GetScore}(H)$ \\
    BestIndex $= \arg\max(S_1, S_2, \dots, S_N)$ \\
    NewSubgoal = $H$[BestIndex] \\
    $S_{new} = [S_1, S_2, \dots, S_N][\text{BestIndex}]$ \\
	\tcp{\textbf{Subgoal Maintenance}}
	OldSubgoal = previous subgoal \\
    OldSubgoal = \texttt{UpdatePos}(OldSubgoal) \\
    $[S_{old}]$ = \texttt{GetScore}([OldSubgoal]) \\
    \eIf{\texttt{GetDist}(OldSubgoal, $p_O$) $< d_1$  \textbf{or} $S_{new} - S_{old} > d_2$}{
        subgoal = NewSubgoal
    }{  
		\tcp{Retain the old subgoal}
        subgoal = OldSubgoal \\
    }
    action = \text{LocalPlanner}(subgoal)
	\label{all_frame}
\end{algorithm}

In LME, the policy is fully represented by an interpretable programming language.
Inspired by previous work~\cite{jang2021hindsight,brito2021go}, LME employs subgoals to decompose the long-horizon decision-making task. The policy framework consists of three modules: a subgoal generation program, a subgoal maintenance program, and a local planner. At each step, the policy follows a sequential process: the subgoal generation program infers a new subgoal based on the local observation and the final target. Then, the subgoal maintenance program compares the new subgoal with the previous one and selects the optimal one as the subgoal. Finally, the local planner generates the appropriate actions based on the selected subgoal. The overall framework is presented in Algorithm~\ref{all_frame}.

\subsection{Subgoal Generation Program}

\begin{figure*}[t]  
    \centering
    \begin{subfigure}[t]{0.70\columnwidth}
        \centering
        \includegraphics[width=0.83\linewidth]{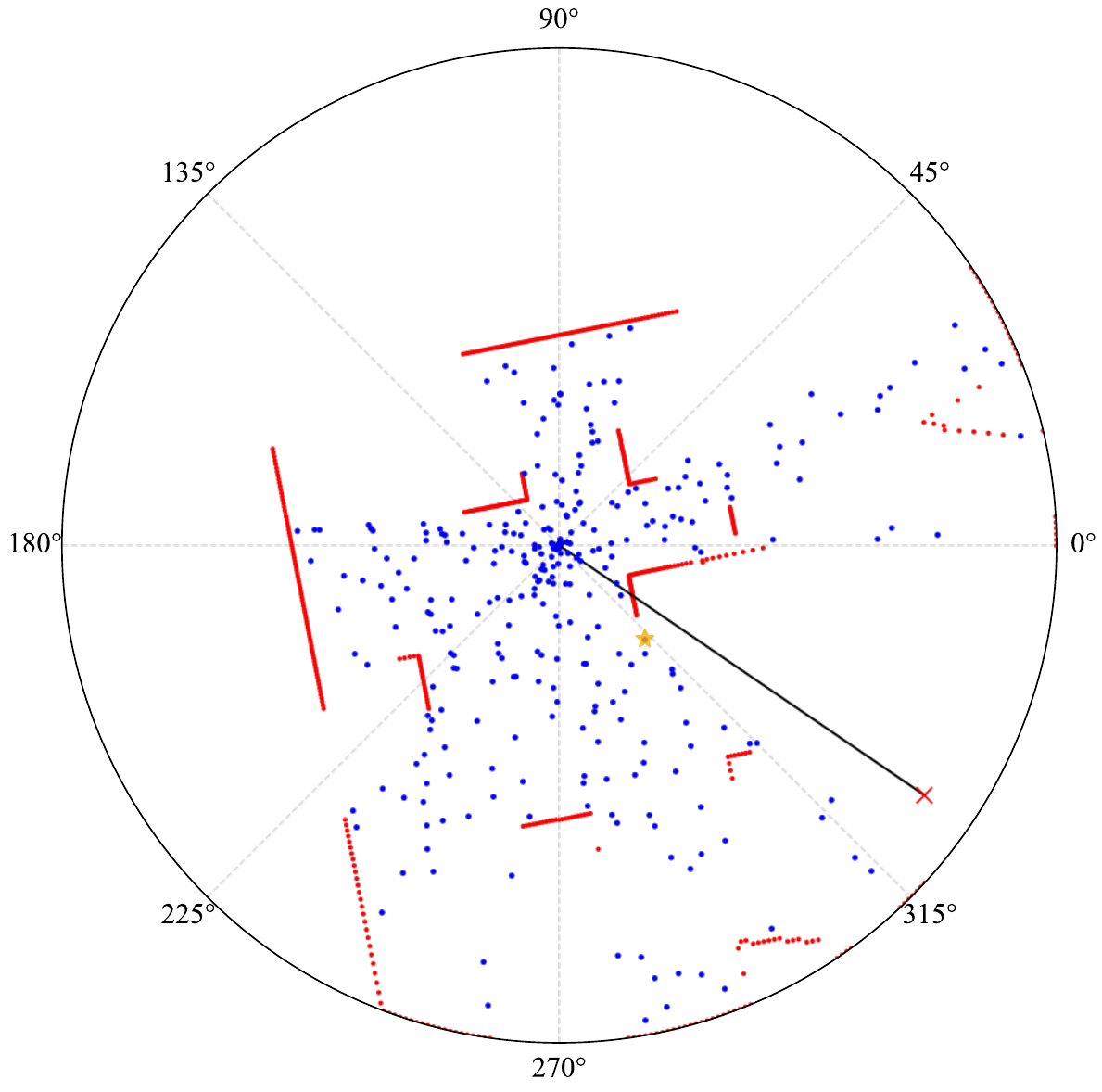}
        \caption{Illustration of candidate points and subgoals. The robot is at the origin, and its orientation is always aligned with the positive y-axis.
	The red points represent scan points (i.e., obstacles), the blue points are the sampled candidate points, the red ``x" marks the navigation target, and the yellow star represents the generated subgoal.
		}
        \label{fig:candidate}
    \end{subfigure}
	\hspace{0.01\columnwidth}
    \begin{subfigure}[t]{0.60\columnwidth}
        \centering
        \includegraphics[width=\linewidth]{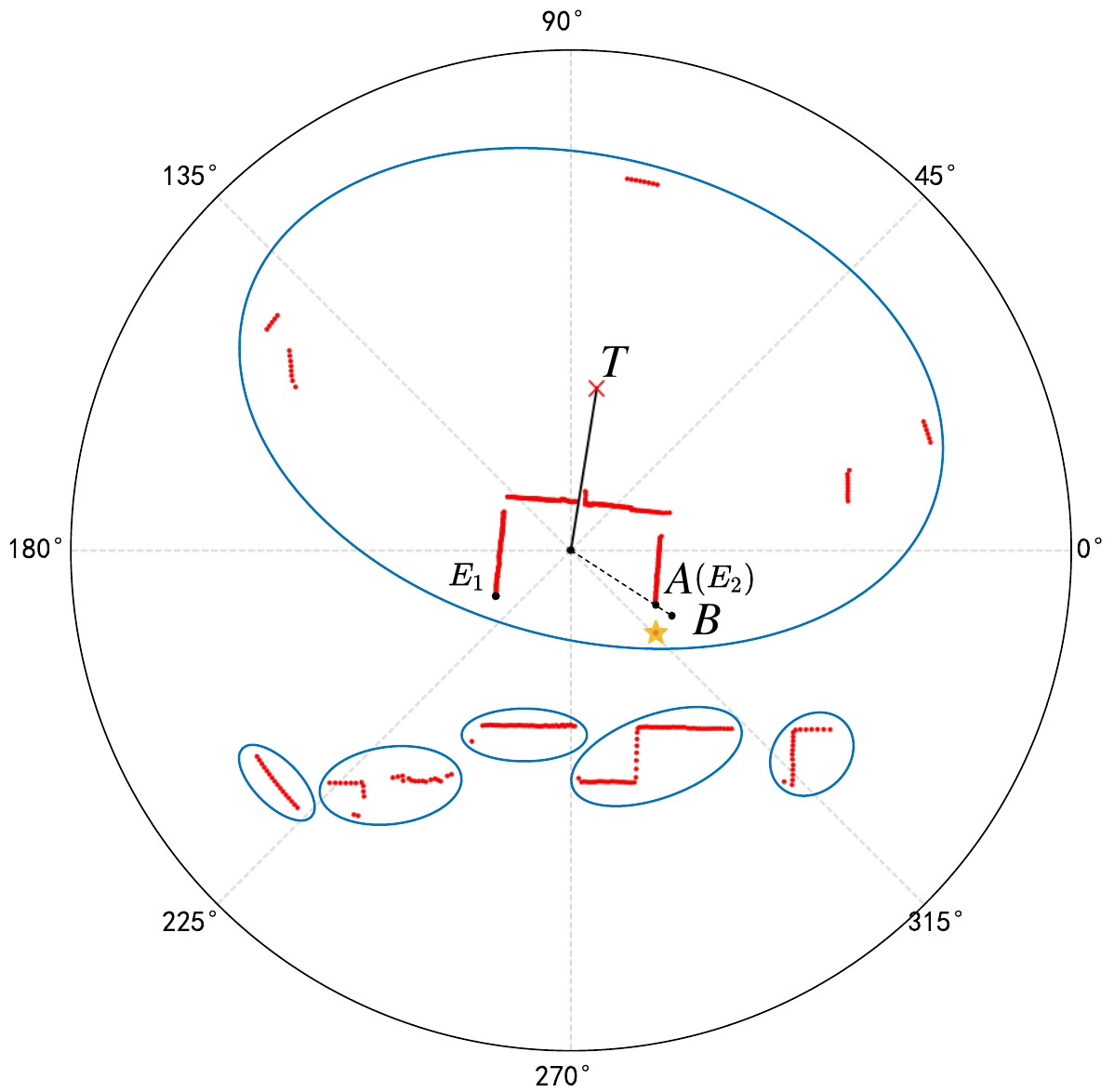}  
        \caption{Illustration of the computation of $a_{\text{mount}}$. The clustering results of the scan points are shown, where red points enclosed by a blue circle are grouped into the same obstacle. Point $B$ lies on the extension of the ray $OA$, and the length of segment $AB$ is $r$.} 
        \label{subfig:left}
    \end{subfigure}
	\hspace{0.01\columnwidth}
    \begin{subfigure}[t]{0.60\columnwidth}
        \centering
        \includegraphics[width=\linewidth]{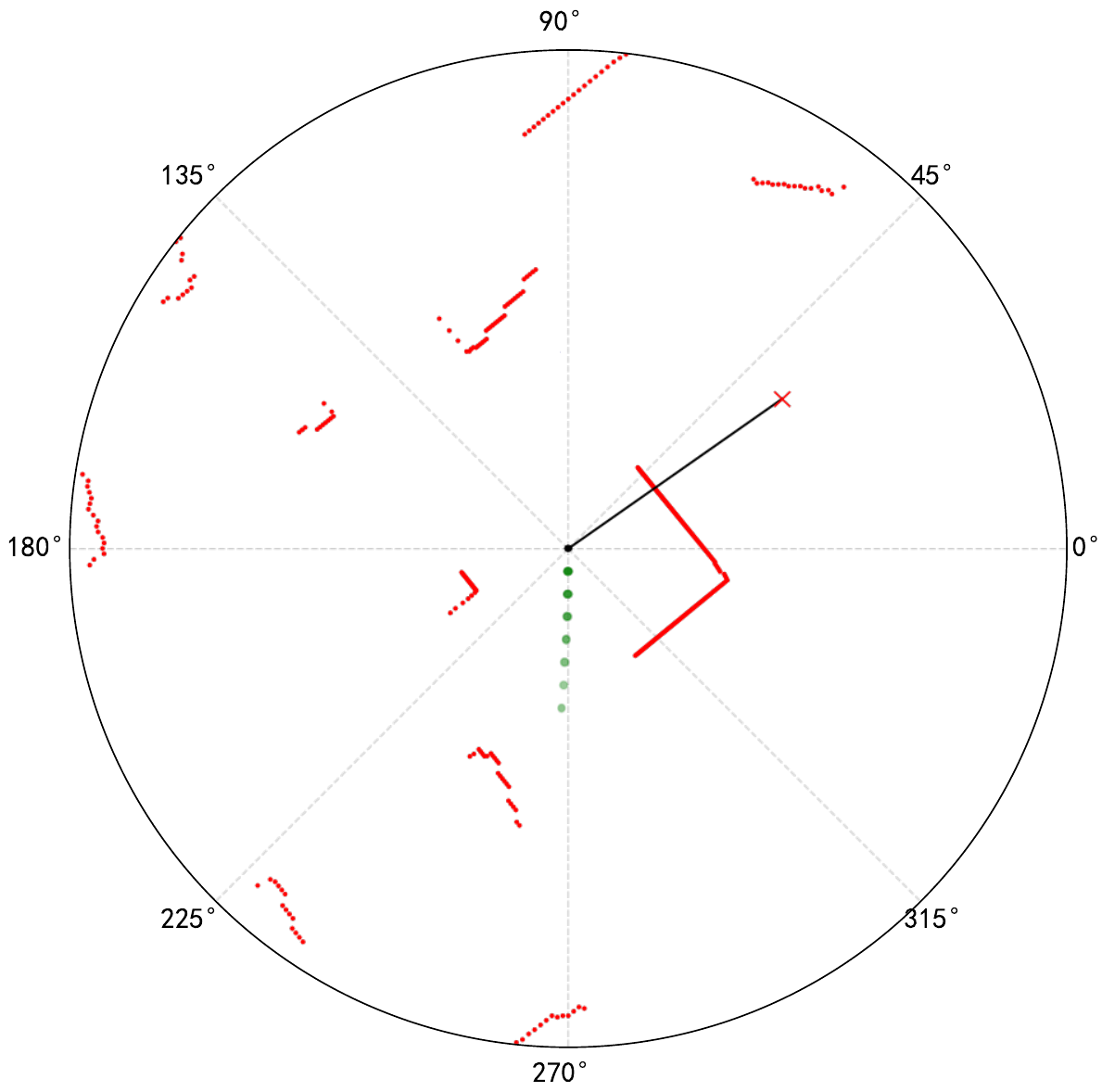}
		\caption{Illustration of historical positions. The green points represent historical positions, where darker colors indicate positions closer to the current time step.}
        \label{subfig:right}
    \end{subfigure}
	\caption{Three separate figures.}
	\vspace{-0.2cm}
\end{figure*}

In complex environments, selecting appropriate subgoals can effectively decompose the task, thereby reducing decision-making complexity. This section introduces our subgoal generation program, which processes 2D LiDAR data to reason about the relative position of the subgoal (i.e., the distance and angle relative to the robot's origin).

To effectively circumvent the challenges associated with directly processing high-dimensional LiDAR observations, LME randomly samples a set of candidate points within the 2D LiDAR sensing range. Each candidate point are then scored by a program, and the point with the highest score is selected as the subgoal. Fig.~\ref{fig:candidate} illustrates the candidate sampling.

To identify effective and feasible subgoals, we define multiple selection criteria, each of which extracts specific features for every candidate point. 

we assume there exists a candidate point $i$, with its relative position to the origin denoted as $p_i$, with Cartesian coordinates $x_p$ and $y_p$. For clarity, we omit the subscript $i$ in the Cartesian coordinates. The relative position of the target point $T$ is denoted as $p_T$. Additionally, we assume that the robot has a circular shape with a \textbf{collision radius} $r$. 

\textbf{(1) Target proximity.} To encourage the robot to approach the target, candidate points that are closer to the target are preferred. We define the following feature:
\begin{equation}
  a_{\text{close}} = -\texttt{GetDist}(p_i, p_T),
\end{equation}
where \texttt{GetDist} calculates the Euclidean distance between two points. As a result, candidate points with smaller distances to the target yield higher values of $a_{\text{close}}$.

\textbf{(2) Directional deviation.} 
Candidate points that deviate significantly from the robot’s forward direction are penalized. This feature discourages frequent heading changes and promotes the selection of subgoals that lie closer to the robot’s frontal direction. We define:

\begin{equation}
a_{\text{dev}} = -\,\left|\operatorname{atan2}\!\left(y_p, x_p\right) - \frac{\pi}{2}\right|.
\end{equation}

\textbf{(3) Collision-free reachability (origin to subgoal).}  
Within the robot’s visible field, the straight-line path from the origin to a candidate point should maintain a clearance of at least $r$ from all obstacles (i.e., scan points). Otherwise, the robot would collide with an obstacle while moving toward the candidate point. We therefore define the following collision feature:
\begin{equation}
\begin{split}
c_{\text{o}} &= \texttt{CheckCollision}(p_O, p_i, \textit{scan}, r), \\
\end{split}
\end{equation}
where \textit{scan} denotes the set of all scan points, and \texttt{CheckCollision} determines whether the line segment from $p_O$ to $p_i$ comes within $r$ meters of any scan point. A point is deemed favorable if $c_{\text{o}}$ is false.

\textbf{(4) Optimistic target reachability (subgoal to target).}  
Within the visible field, the straight-line path from the candidate point to the target is required to maintain a clearance of more than $r$ from all obstacles. This feature provides an optimistic estimate by assuming that no obstacles exist in unobserved regions, thereby encouraging the selection of subgoals that can directly reach the target while bypassing obstacles ahead. The corresponding feature is defined as:
\begin{equation}
\begin{split}
c_{\text{t}} &= \texttt{CheckCollision}(p_i, p_T, \textit{scan}, r). \\
\end{split}
\end{equation}

\begin{algorithm}[t]
\caption{Computation of the $a_{\text{mount}}$}
Cluster the LiDAR scan points into multiple sectors. \\
Identify the sector that blocks the direct line of sight to the target, and denote the scan points in this sector as $\textit{scan}_b$. \\

\eIf{\texttt{CheckCollision}$(p_i, p_T, \textit{scan}_b, r)$}{
    Determine the boundary point $A$ of the blocking sector that is closest to the target direction. \\
    Generate point $B$ by moving $r$ meters from point $A$ along the ray $OA$. \\
    $a_{\text{mount}} = -\,\texttt{GetDist}(p_i, p_B)$
}{
    $a_{\text{mount}} = 0$
}
\label{alg:mount}
\vspace{-0.2cm}
\end{algorithm}

\textbf{(5) Surmount feature (escaping local optima).}  
The surmount feature is the key criterion for enabling the robot to surmount local optima regions. This feature explicitly encourages behaviors that bypass obstacles responsible for trapping the robot in locally optimal.

To compute this feature, we first cluster the LiDAR scan points. For any pair of scan points, if their Euclidean distance is less than $2r$, the robot cannot pass between them. Consequently, scan points separated by less than $2r$ are considered to belong to the same obstacle, and all scan points lying between them are grouped accordingly. As a result, the scan points over the full $360^\circ$ field of view can be partitioned into multiple \emph{sectors}, where the scan points in each sector represent an obstacle that the robot cannot traverse directly, as illustrated in Fig.~\ref{subfig:left}. The time complexity of clustering $M$ scan points is $O(M^2)$. However, by employing efficient spatial data structures such as KD-trees~\cite{bentley1975multidimensional}, the complexity can be reduced to $O(M \log M)$.

When the target lies behind a particular sector, moving directly toward the target may temporarily reduce the Euclidean distance but will not lead to reaching the goal, thereby causing the robot to become trapped in a local optimum. In such cases, the appropriate strategy is to bypass the blocking sector rather than heading straight toward the target.

Each sector has two boundary scan points, defined as the points with the minimum and maximum angular values within the sector, denoted by $E_1$ and $E_2$, respectively. We compute the angles $\angle E_1 O T$ and $\angle E_2 O T$, and select the boundary point with the smaller angle relative to the target, referred to as point $A$. Subsequently, a point $B$ is generated along the line segment $OA$ at a distance of $r$ from point $A$. Candidate points that are closer to point $B$ are assigned higher surmount values. The complete procedure for computing the surmount feature is summarized in Algorithm~\ref{alg:mount}.

\textbf{(6) Historical avoidance.}  
To prevent the agent from moving backward and degrading navigation efficiency, we introduce an additional criterion that penalizes candidate points whose straight-line paths intersect previously visited regions.

As illustrated in Fig.~\ref{subfig:right}, we maintain a first-in-first-out (FIFO) queue that stores the robot’s positions over the most recent $k$ time steps. These historical positions are treated as virtual obstacles, denoted by $P_{\text{his}}$. The straight-line path from the origin to a candidate point is required to maintain a clearance of at least $r$ from all points in $P_{\text{his}}$. The corresponding feature is defined as:
\begin{equation}
\begin{split}
c_{\text{h}} &= \texttt{CheckCollision}(p_O, p_i, P_{\text{his}}, r). \\
\end{split}
\end{equation}
The historical positions are updated at each time step based on the robot’s linear and angular velocities, as described in subsection \ref{maintain}. In this study, we set $k=8$, which strikes a balance between retaining sufficient historical context and avoiding an overly large penalty region that would unnecessarily restrict movement.

The score $S$ is defined as the weighted sum of all features:
\begin{equation}
\begin{split}
S ={}& w_1 a_{\text{close}} + w_2 a_{\text{dev}} + w_3 I(c_{\text{o}}) \\
     &+ w_4 I(c_{\text{t}}) + w_5 a_{\text{mount}} + w_6 I(c_{\text{h}}),
\end{split}
\end{equation}
where $I(\cdot)$ is an indicator function, and $w_1$ through $w_6$ denote the learnable weights associated with each feature. Nevertheless, manually adjusting and coordinating these weights is no trivial task. How to automatically optimize these parameters will be elaborated in the following sections.

\subsection{Subgoal Maintenance Program}
\label{maintain}

The subgoal generation module introduced in the previous subsection infers a new subgoal at every decision step. However, frequently switching subgoals can degrade navigation efficiency, while persistently following an old subgoal until it is reached may miss better alternatives. To address this trade-off, we introduce a subgoal maintenance module, which dynamically decides whether to retain the previous subgoal or switch to a newly generated one. As shown in the second part of Algorithm~\ref{all_frame}, the subgoal maintenance program determines the final subgoal by checking whether the previous subgoal has been reached and by comparing the scores of the two subgoals. The parameters $d_1$ and $d_2$ in the algorithm are subject to optimization.

At each decision step, the score of the previous subgoal is recomputed using the current observation. Since absolute coordinates are not available, the relative position of the previous subgoal should also be updated accordingly. The update rule (\texttt{UpdatePos} in Algorithm~\ref{all_frame}) proceeds as follows:

Given a previous subgoal in polar coordinates $(d, \theta)$, we convert it to Cartesian coordinates $(x, y)$. 
The forward displacement over one control period $\Delta t$ is $\Delta y = \frac{v_{\text{prev}} + v}{2} \cdot \Delta t$, where $v_{\text{prev}}$ and $v$ are the previous and current linear velocities. 
We translate the subgoal backward along the forward direction: $y \leftarrow y - \Delta y$, then convert back to polar coordinates $(d', \theta')$. 
Finally, we compensate for rotation: $\theta' \leftarrow \theta' - \frac{\omega_{\text{prev}} + \omega}{2} \cdot \Delta t$, with $\omega_{\text{prev}}, \omega$ being the previous and current angular velocities.

This update approximates the subgoal's relative motion in the robot-centric frame using a first-order motion model combining forward translation and rotational compensation.


\subsection{Local Planner}

\begin{algorithm}[t]
\caption{Low-Level Velocity Control}
\label{velocity_control}
\KwIn{Current robot velocity $(v, \omega)$, subgoal $(d, \theta)$}
\KwOut{Velocity commands $(v_{rate}, \omega_{rate})$}
\eIf{$|\theta - \pi/2| < 9^\circ$}{
    Compute forward velocity $v_{rate}$ using a linear acceleration-limited controller.
}{
    $v_{rate} \leftarrow 0$
}
Compute angular velocity $\omega_{rate}$ using an acceleration-limited angular controller. \\
\Return $(v_{rate}, \omega_{rate})$
\end{algorithm}

Given a subgoal specified by its relative distance and bearing, the local planner directly outputs control actions based on the robot’s current velocity and its maximum velocity and acceleration constraints. Since the previously described program generates subgoals that are reasonable and geometrically reachable, the decision-making burden of the local planner is significantly reduced.

In this subsection, we present a simple local planner. Importantly, LME does not impose any restrictions on the type of low-level controller used. In principle, any existing local planner can be integrated, including classical methods such as DWA~\cite{fox_dynamic_1997} or DRL-based approaches~\cite{zhang2025drl,xie_drl-vo_2023}. In the experimental section, we further explore the use of alternative local planners to demonstrate this flexibility.

Algorithm~\ref{velocity_control} outlines the structure of the local planner. Linear and angular velocities are computed by two independent controllers. The angular velocity controller continuously steers the robot toward the direction of the subgoal. The linear velocity controller outputs a positive forward velocity only when the subgoal lies in front of the robot; otherwise, the linear velocity is set to zero to avoid motions.

Both the linear and angular controllers adopt a simple acceleration-limited control scheme. Specifically, when the robot is far from the desired state, the controller accelerates with the maximum allowable acceleration; as it approaches the target, it decelerates symmetrically to ensure that the velocity approaches zero at convergence. 
When the remaining distance exceeds the stopping distance, the controller maintains a constant velocity. 
In addition, the angular velocity controller switches to a PD control behavior when the angular error becomes small, which helps reduce oscillations near the target direction.

\subsection{Program Optimization}
\label{optimization}

The program contains several parameters (including $w$ and $d$) that need to be optimized. However, due to the discrete structure of the program, these parameters are non-differentiable and therefore cannot be optimized via gradient-based methods. Following prior program optimization work~\cite{verma2018programmatically,gu2025pi}, we employ gradient-free Bayesian optimization~\cite{snoek2012practical} to determine the optimal values of all parameters. The optimization objective is to maximize the reward $R$.
Parameters $w$ and $d$ are constrained to be positive real numbers. The number of optimization steps is set to 150.

\subsection{Extension to Non-Circular Robot Shapes}

In the previous sections, for clarity of presentation, we assumed that the robot has a circular footprint.
For a circular robot, the collision radius is identical in all directions. However, for robots with non-circular footprints, such as square or rectangular shapes, the collision radius varies with direction. In this subsection, we discuss how LME can be extended to handle robot shapes beyond circular footprints.

Let $r_{\max}$ and $r_{\min}$ denote the maximum and minimum collision radii of the robot, over all orientations, respectively. If the collision radius $r$ used in the subgoal generation module satisfies $r \geq r_{\max}$, any selected subgoal is guaranteed to be reachable from all orientations. Conversely. If $r < r_{\min}$, the subgoal is unreachable. When $r \in [r_{\min}, r_{\max}]$, reachability depends on the robot’s orientation and is therefore restricted to a subset of feasible orientations.

Given that the local planner introduced in the previous subsection is designed under simplified assumptions, we adopt a conservative strategy by setting the collision radius to $r = r_{\max}$. Although this choice may exclude certain subgoals that are reachable, it guarantees collision safety.

Finally, a notable advantage of LME is that adapting to a different robot shape only requires updating the single radius parameter (i.e., setting $r$). No policy retraining is needed.

\begin{table}[thp]
	\centering
	\caption{The performance of LME in ROS Stage.}
	\begin{tabular}{@{}c|ccc@{}}
		\toprule
		Environment & \textbf{Success rate} $\uparrow$ & \textbf{Collision rate} $\downarrow$ & \textbf{Timeout rate} $\downarrow$ \\ 
		\midrule
		Env1 & 0.95±0.01 & \textbf{0.04}±0.01 & \textbf{0.02}±0.01 \\  
		Env2 & 0.94±0.01 & \textbf{0.05}±0.02 & \textbf{0.00}±0.01 \\  
		\bottomrule
	\end{tabular}
	\label{stage}
	\vspace{-0.2cm}
\end{table}

\section{Experiments in Simulation}

\subsection{Training in Simulation}

\begin{figure}[]
    \centering
    \begin{subfigure}[t]{0.42\columnwidth}
        \centering
        \includegraphics[width=\linewidth]{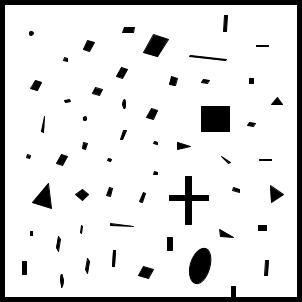}
        \caption{Env1~\cite{zhang2022ipaprec}}
		\label{stage1}
    \end{subfigure}
	\hspace{0.03\columnwidth}
    \begin{subfigure}[t]{0.42\columnwidth}
        \centering
        \includegraphics[width=\linewidth]{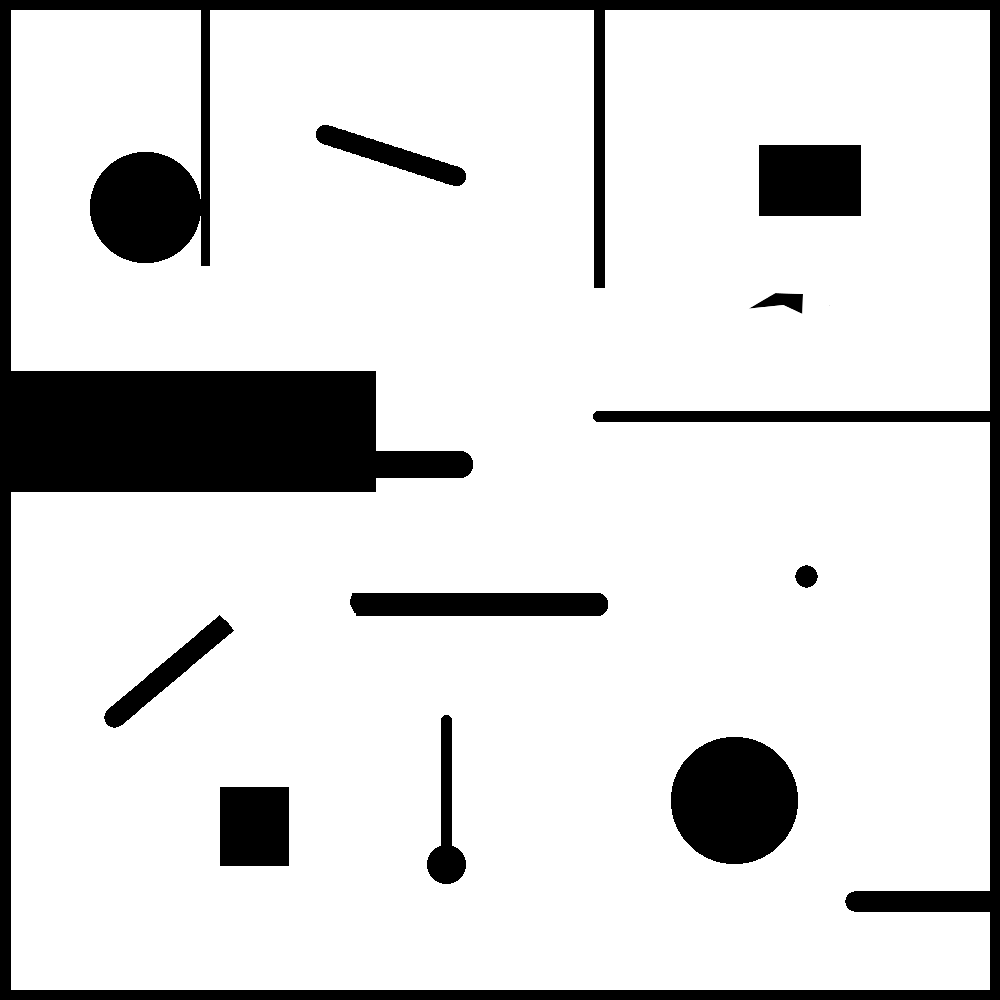}
        \caption{Env2~\cite{pfeiffer2018reinforced}}
		\label{stage2}
    \end{subfigure}
    \caption{Training map and testing map in ROS Stage.}
	\vspace{-0.3cm}
\end{figure}

We use the lightweight ROS Stage simulator~\cite{Stage_rosWiki} to train the policy of LME, and adopt the map used in prior work~\cite{zhang2022ipaprec} as the training environment. The layout of the map is shown in Fig.~\ref{stage1}. The map contains relatively dense small obstacles but does not include large obstacles (e.g., walls); therefore, it contains few local minima. In ROS Stage, the robot control frequency is set to 5 Hz. The LiDAR has a 360-degree field of view, a sensing range of 15 meters, and provides 540 uniformly distributed laser beams. The robot is modeled as a circular shape with a radius of 0.2 m.

In this environment, each episode specifies an initial robot position and a goal position (we refer to these two positions together as a configuration). Before training begins, LME randomly samples several configurations, which remain fixed throughout training and are used to compute the program evaluation metric $R$. This design ensures that the evaluation metrics of different programs are comparable. The training process consists of a total of 600 episodes. All experiments are conducted on a desktop computer running Ubuntu 20.04 with an Intel i9-13900KF CPU, using a Python-based implementation. Under this setup, a single training run of LME requires approximately 1.5 hours to complete.


We report the navigation performance of LME on both the training map and an unseen map (i.e., Fig.~\ref{stage2}). The unseen map contains relatively long walls. When the goal is located on the opposite side of a wall, the robot must navigate around the wall to reach the destination, which introduces several local minimum regions. On each map, we randomly sample 50 configurations to evaluate policy performance. The experimental results are summarized in Table~\ref{stage}. As can be seen, our method achieves success rates close to 95\% on both maps. Even though the layouts of Env2 and Env1 differ significantly, the policy trained in Env1 transfers to Env2 without noticeable performance degradation.

Unless otherwise specified, all subsequent simulation experiments and real-world experiments use the LME policy learned in ROS Stage.

\subsection{Performance Validation in BARN}

\begin{figure}[t]
	\centering
	\includegraphics[width=0.66\columnwidth]{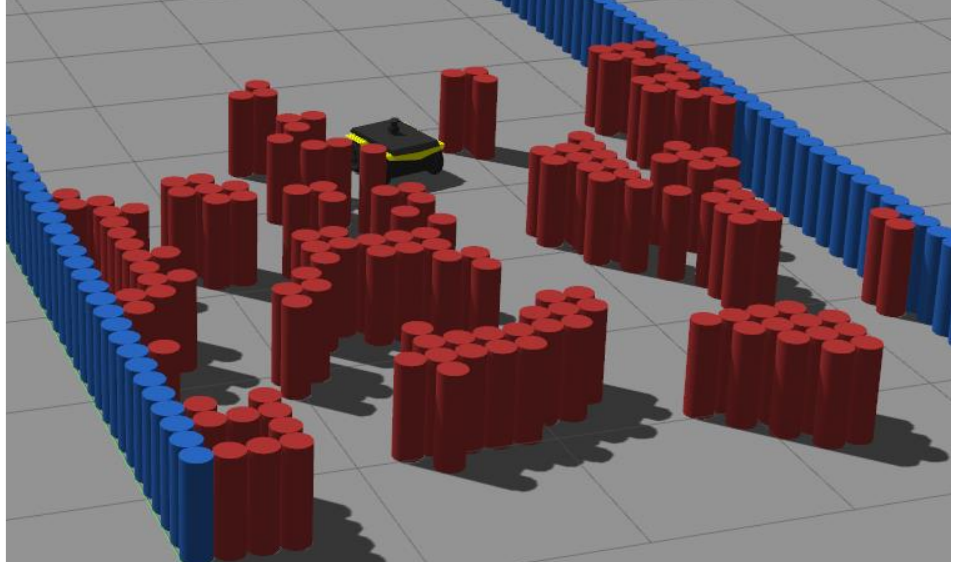}
	\caption{A BARN scenario example.}
	\label{barn_fig}
\end{figure}

\begin{table*}
	\centering
	\caption{The performance of all methods on BARN environment.}
	\setlength{\tabcolsep}{8pt}
	\begin{tabular}{@{}c|cccc|c@{}}
		\toprule
		\toprule
		& \textbf{Success rate} $\uparrow$ & \textbf{Collision rate} $\downarrow$ & \textbf{Timeout rate} $\downarrow$ & \textbf{BARN Metric} $\uparrow$ & \textbf{Use global planner} \\ \midrule
		DWA         & 0.607±0.0094          & 0.253±0.0236        & 0.140±0.0141      & 0.3015±0.0049     & No \\
		DRL-DCLP 	 & 0.713±0.0094         & 0.077±0.0249      & 0.210±0.0163      & 0.3475±0.0052    & No \\
		S$^3$-FISVFH  & 0.820±0.0216        & 0.113±0.0189      & 0.067±0.0047      & 0.3812±0.0074   	& No \\ \midrule
		E-Band   & 0.653     & 0.337         & 0.01  & 0.288 & \textbf{Yes} \\
		KUL+FM 	 & 0.860     & \textbf{0.02} & 0.12   & 0.388 & \textbf{Yes} \\
		DRL-VO  &  0.810   & 0.190 & \textbf{0} & 0.252 & \textbf{Yes} \\
		DRL-DCLP & 0.880*  & 0.063* & 0.057 & 0.388 & \textbf{Yes}   \\  \midrule

		LME-DWA   & 0.860±0.0100    & 0.130±0.0100      & 0.010*±0.0000    & \textbf{0.419}±0.0049      & No  \\  
		LME-DRL  & 0.860±0.0100    & 0.095±0.0050    & 0.045±0.0150   & 0.403±0.0005	& No \\  
		LME    & \textbf{0.883}±0.0094  & 0.100±0.0216 & 0.017±0.0125   & \textbf{0.419}±0.0084  & No  \\  \bottomrule
	\end{tabular}
	\\[0.2cm]  
    \begin{minipage}{0.78\linewidth}
    \textbf{Note:}
values are presented as mean ± standard deviation. The second best methods are denoted with *. The results of the three methods using global planners are directly taken from DRL-DCLP's original paper~\cite{zhang2025drl}.
    \end{minipage}
	\label{barn_all}
	\vspace{-0.2cm}
\end{table*}

\begin{figure*}[t]
	\centering
	\includegraphics[width=1.8\columnwidth]{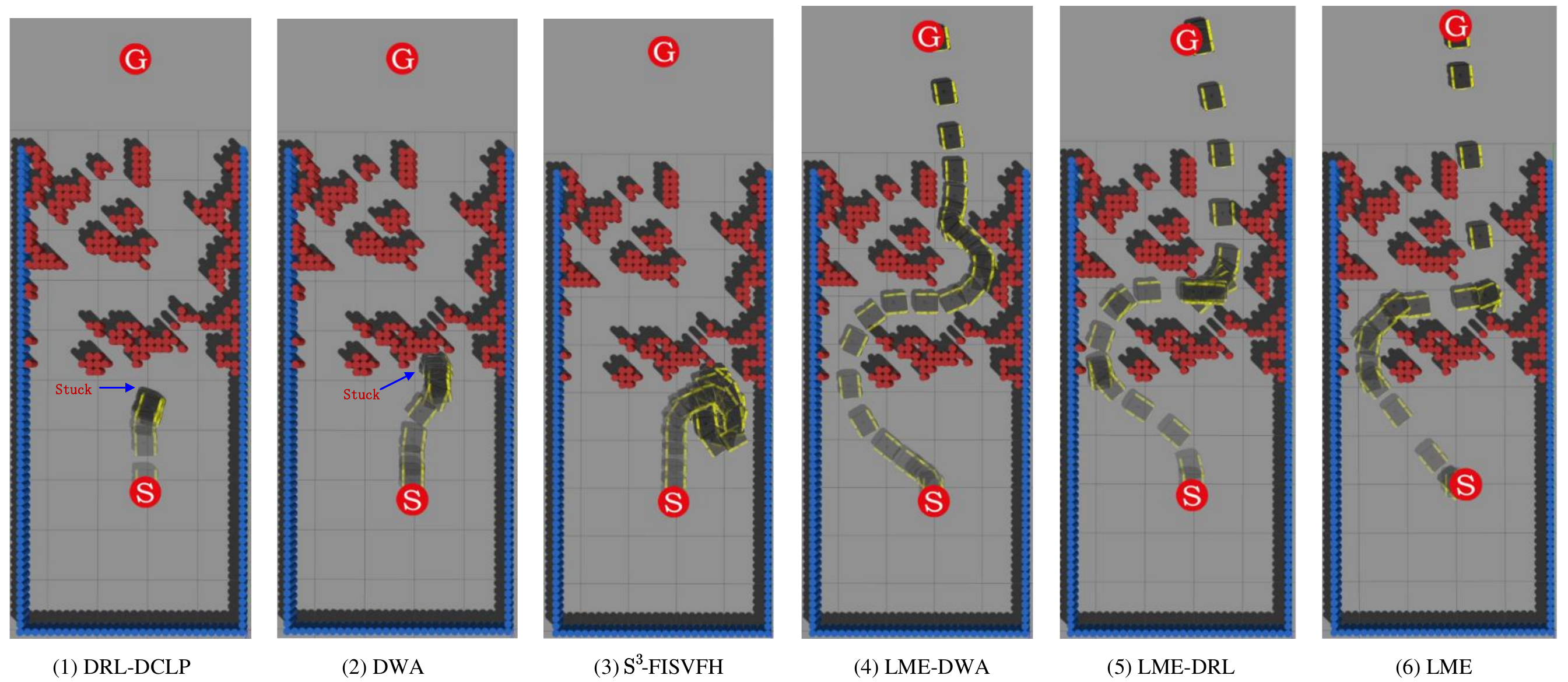}
	\caption{Trajectories of several methods in BARN. The corresponding videos
are provided in the supplementary file.} 
	\label{barn_trace}
	\vspace{-0.2cm}
\end{figure*}

We select the Benchmark for Autonomous Robot Navigation (BARN) Challenge~\cite{xiao2023autonomous,xiaoautonomous} as the evaluation environment to further assess the performance of LME. BARN is a mobile robot navigation challenge hosted at the ICRA conference and has been held annually for four consecutive years. As illustrated in Fig.~\ref{barn_fig}, participants are required to navigate robots through highly constrained environments, which also contain regions prone to trapping the robot in local optima. For large-scale evaluation, we follow prior work~\cite{zhang2025drl} and select the 100 challenging maps from the BARN benchmark to assess the performance of each method.

In the BARN challenge, the controllable robot is the Jackal robot, a four-wheel differential-drive mobile platform. The robot has a rectangular shape with a length of 0.42 m and a width of 0.31 m, and its maximum linear velocity is 2m/s. The evaluation metric used in the BARN Challenge is defined as follows for each testing map:
\begin{equation}
\begin{split}
	\text{BARN Metric} = I(\text{success}) \times \frac{OT}{\text{clip}(AT, 2OT, 8OT)},
\end{split}
\end{equation}
where $I(\text{success})$ is an indicator function that equals 1 if the robot successfully reaches the goal without collision, and 0 otherwise. $AT$ denotes the actual traversal time, and $OT$ denotes the optimal traversal time. The optimal traversal time $OT$ is computed as the length of the shortest path divided by the robot’s maximum speed.

To conduct a comprehensive comparison, we evaluate LME against several classical and state-of-the-art DRL approaches:
\begin{itemize}
\item Dynamic Window Approach (DWA)~\cite{fox_dynamic_1997}.
For DWA, we do not use the official BARN implementation, as it invokes a global planner by default. Instead, we manually implement the DWA algorithm and tune its parameters specifically for the BARN environment.

\item DRL-DCLP~\cite{zhang2025drl}. DRL-DCLP is a recently proposed method that can control robots of arbitrary sizes in a zero-shot manner. Since it reports competitive performance on the BARN benchmark, we include it in our comparison.

\item Search-Smooth-Safeguard Fuzzy Inference System Vector Field Histogram (S$^3$-FISVFH)~\cite{xiaoautonomous}.
This method was proposed by the SSRL team and won first place in the 2025 BARN Challenge. It builds upon FISVFH~\cite{balan2019fuzzy} by optimizing the internal fuzzy controller parameters and integrating A* search, smoothing filters, and a safeguard mechanism to enhance the original FISVFH method. 

\item LME-DWA and LME-DRL.
Since the subgoals inferred‌ by LME are agnostic to the underlying local planner, LME can be readily combined with different low-level controllers. Specifically, we adopt DWA and DRL-DCLP as the low-level controllers, resulting in two variants, termed LME-DWA and LME-DRL, respectively. Notably, both controllers can be seamlessly integrated into LME without any retraining or additional parameter tuning.

\item Methods with global planners.
In addition to the above methods, we also report results from approaches that employ global planners, such as Elastic Bands (E-Band)~\cite{quinlanElasticBandsConnecting1993}, KUL+FM~\cite{xiao2023autonomous}, and DRL-VO~\cite{xie_drl-vo_2023}. These methods leverage a global path generated by a classical global planner (i.e., Navfn-ROS~\cite{Navfn}), which provides access to long-horizon information beyond local observations. As a result, they benefit from a advantage compared to mapless navigation approaches.

\end{itemize}

The experimental results of all methods are summarized in Table~\ref{barn_all}. Our method achieves the best overall performance, consistently outperforming both traditional and DRL-based approaches, demonstrating the effectiveness of LME for navigation.
A comparison between DWA and DRL-DCLP and their LME-augmented counterparts (i.e., LME-DWA and LME-DRL) further highlights the benefits of LME-generated subgoals. In particular, incorporating LME substantially improves the performance of both planners, providing strong evidence for the effectiveness of the proposed subgoal generation and maintenance mechanisms. For DWA, the success rate increases from 0.60 to 0.85, accompanied by substantial reductions in both collision and timeout rates. Similarly, the DRL-based methods achieve higher success rates while exhibiting a pronounced reduction in timeout frequency.
Finally, compared with methods that rely on global planners, LME achieves a slightly higher success rate while operating solely on local observations. Moreover, LME-DWA and LME-DRL attain performance comparable to that of approaches equipped with global planners, demonstrating that LME can effectively compensate for the lack of global information through local subgoal generation.

\subsection{Trajectory Visualization}

To visually compare the behavioral patterns of various methods, we present their trajectories on a relatively challenging map. As shown in Fig.~\ref{barn_trace}, LME is able to efficiently navigate around complex obstacles. This strongly demonstrates that LME is capable of overcoming challenging local minima.

Both DWA and DRL-DCLP get stuck in local minima. This is primarily due to their short-sightedness, as they focus solely on short-term rewards or costs and lack guarantees from a global mechanism. However, when combined with our approach, DWA and DRL-DCLP can also easily overcome these local minima.

S$^3$-FISVFH, when faced with a dense cluster of obstacles ahead, continues to loop in circles, eventually entering an infinite loop. Although S$^3$-FISVFH incorporates mechanisms such as forward search to avoid deadlocks, this approach does not take into account the global distribution of obstacles and may not always escape local optima.

\subsection{Interpretability of LME}

This subsection examines the interpretability of our policy. We first report the optimized parameters of the policy. The weights $w_1$ through $w_6$ are optimized to 1.51, 1.56, 999.8, 205.84, 88.7, and 134.9, respectively, while the parameters of the subgoal maintenance module, $d_1$ and $d_2$, are set to 0.51 and 200, respectively.

\begin{figure*}[t]
	\centering
	\includegraphics[width=2\columnwidth]{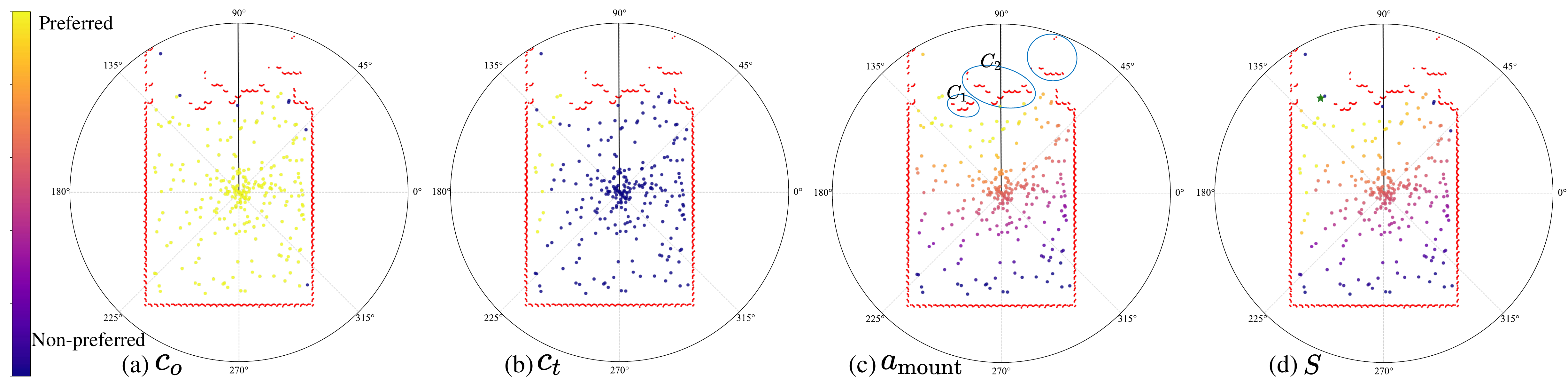}
	\caption{Visualization of features and the score $S$ in BARN. To enhance contrast, extreme values are clipped to the 5th and 95th percentiles. Consistent with previous figures, the black line indicates the target direction, the green star denotes the selected subgoal, and the blue contour highlights a cluster of scan points when calculating $a_{\text{mount}}$.}
	\label{mid1}
\end{figure*}

\begin{figure*}[h]
	\centering
	\includegraphics[width=2\columnwidth]{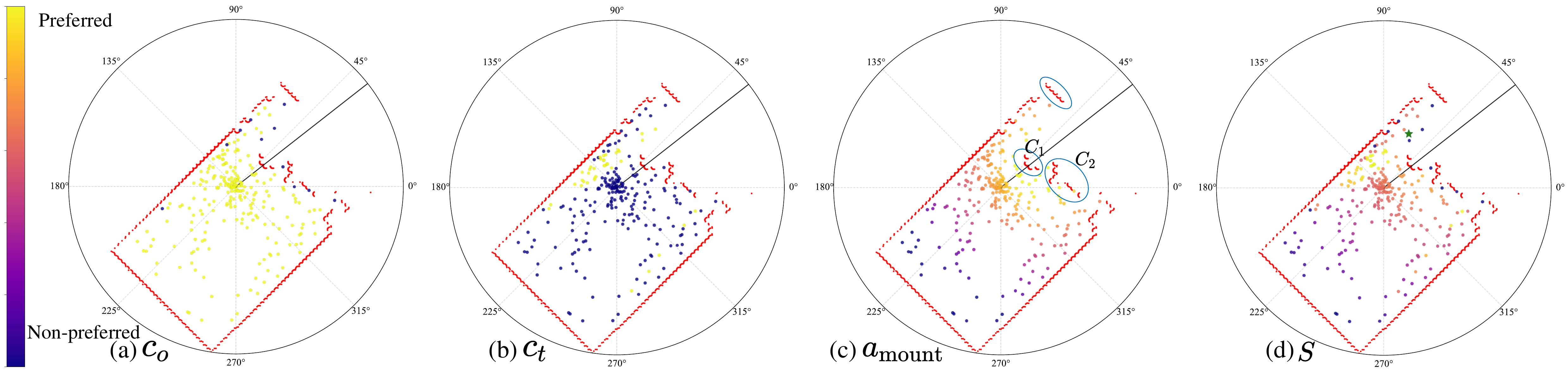}
	\caption{Visualization of features and the score $S$ in BARN.}
	\label{mid2}
\end{figure*}

To further examine the decision-making process of the programmatic policy, we visualize the values of different features for candidate points. Due to space limitations, we present only a subset of the features in Fig.~\ref{mid1} and Fig.~\ref{mid2}. Several observations can be made. When the straight-line path from a candidate point to the origin intersects an obstacle, $c_{\text{o}}$ assigns a low score to that candidate. In contrast, $c_{\text{t}}$ assumes that no obstacles exist in unobserved regions and therefore favors candidate points that appear to provide a direct route toward the target.

Fig.~\ref{mid1}d and Fig.~\ref{mid2}d visualize the distribution of $S$ across all candidate points. Notably, the features exhibit strong complementary effects. Taking Fig.~\ref{mid2} as an example, candidate points near cluster $C_1$ receive high values of $a_{\text{mount}}$. However, some of these candidates still fail to provide a feasible route around the obstacle. By jointly considering multiple features, particularly $c_{\text{t}}$, the policy selects the subgoal marked by the green star in the rightmost panel, which successfully guides the robot around the entire obstacle. In contrast, relying solely on $c_{\text{t}}$ would favor suboptimal candidate points.
	·

This analysis demonstrates that the decision-making process of our programmatic policy is highly interpretable: each decision can be decomposed into the contributions of individual features, allowing its underlying rationale to be explicitly examined. This property provides a clear advantage in interpretability over many DRL-based approaches~\cite{Wang2023,xie_drl-vo_2023}.

\begin{figure}[t]
	\centering
	\includegraphics[width=1.0\columnwidth]{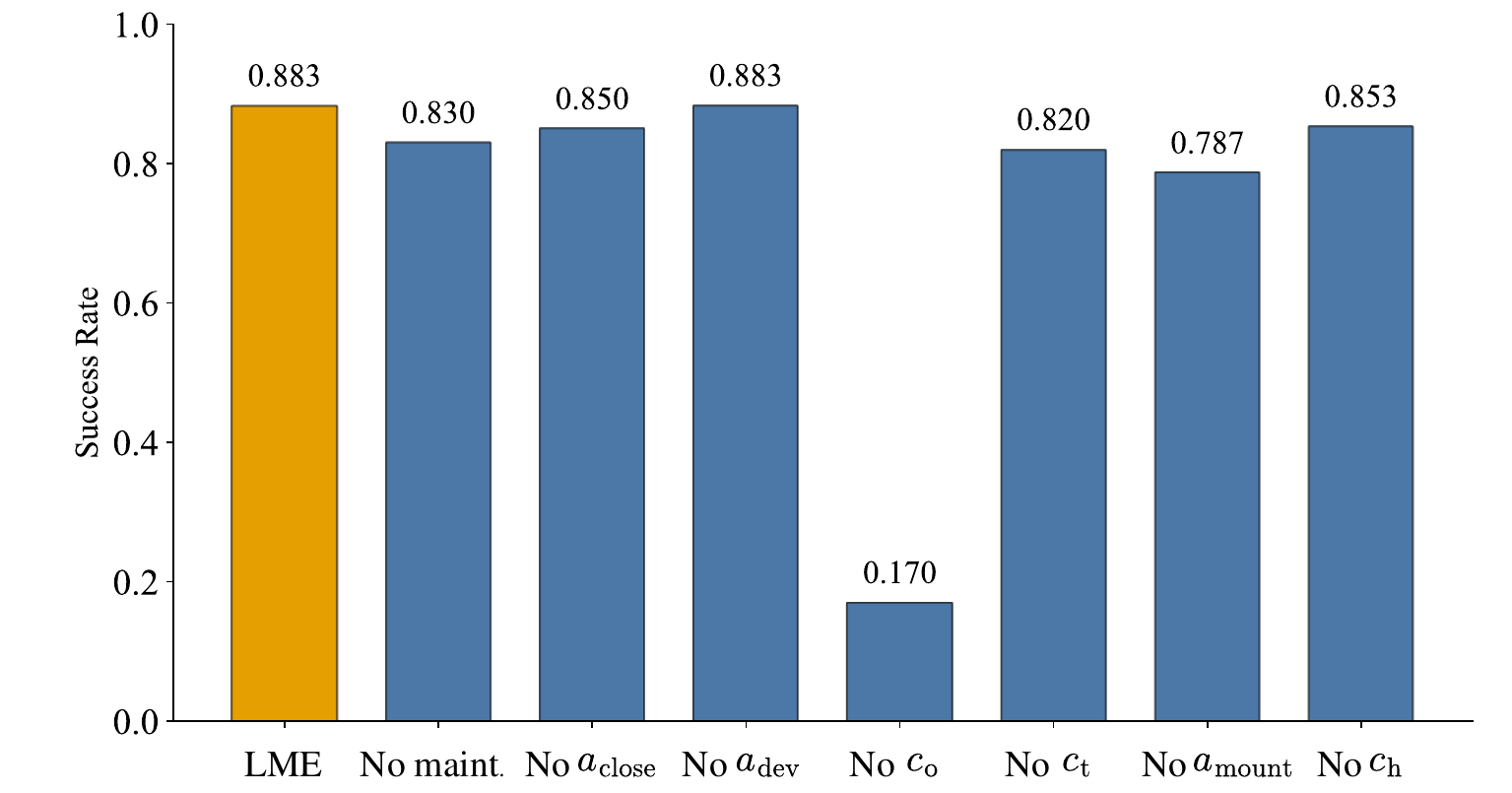}
	\caption{Results of the ablation studies. The success rate is averaged over three independent runs.}
	\label{abl}
\end{figure}

\subsection{Ablation Studies}

In this subsection, we conduct a series of ablation studies in the BARN environment, using the same experimental settings as in the previous evaluations. The subgoal generation module consists of multiple criteria for selecting subgoals. To assess the contribution of each criterion, we remove one score at a time while keeping all others unchanged. Each resulting variant is denoted as No $\cdot$, e.g., No $a_{\text{close}}$. In addition, the subgoal maintenance module adaptively determines whether to retain the previous subgoal. To evaluate its effectiveness, we remove the subgoal maintenance mechanism in Algorithm~\ref{all_frame} and always select the newly generated subgoal at each step, regardless of the previous one. This variant is referred to as No maintenance.

Fig.~\ref{abl} summarizes the results of the ablation studies. By comparing each variant with the complete method, several observations can be made.
First, the collision avoidance score $c_{\text{o}}$ has the most significant impact on navigation performance. Removing $c_{\text{o}}$ results in a substantial decrease in the success rate. 
Second, both $c_{\text{t}}$ and $a_{\text{mount}}$ contribute to the robot's ability to navigate around challenging obstacles. Removing either criterion leads to a noticeable performance degradation.
The subgoal maintenance module also plays an important role in maintaining navigation stability. Without this mechanism, the robot tends to switch subgoals more frequently, resulting in less stable behavior and a lower success rate. 
Furthermore, $a_{\text{close}}$ and $c_{\text{h}}$ both contribute positively to the overall navigation performance, as their removal leads to moderate performance degradation. In contrast, removing $a_{\text{dev}}$ has almost no effect on the success rate. Nevertheless, $a_{\text{dev}}$ encourages the selection of forward-facing subgoals, which is expected to produce more consistent trajectories.
Finally, it should be noted that the scoring criteria are not necessarily independent and may complement each other under different navigation scenarios. Therefore, removing a single criterion does not always result in a substantial performance drop.

\begin{figure*}[t]  
    \centering
    \begin{subfigure}[t]{0.60\columnwidth}
        \centering
        \includegraphics[width=\linewidth]{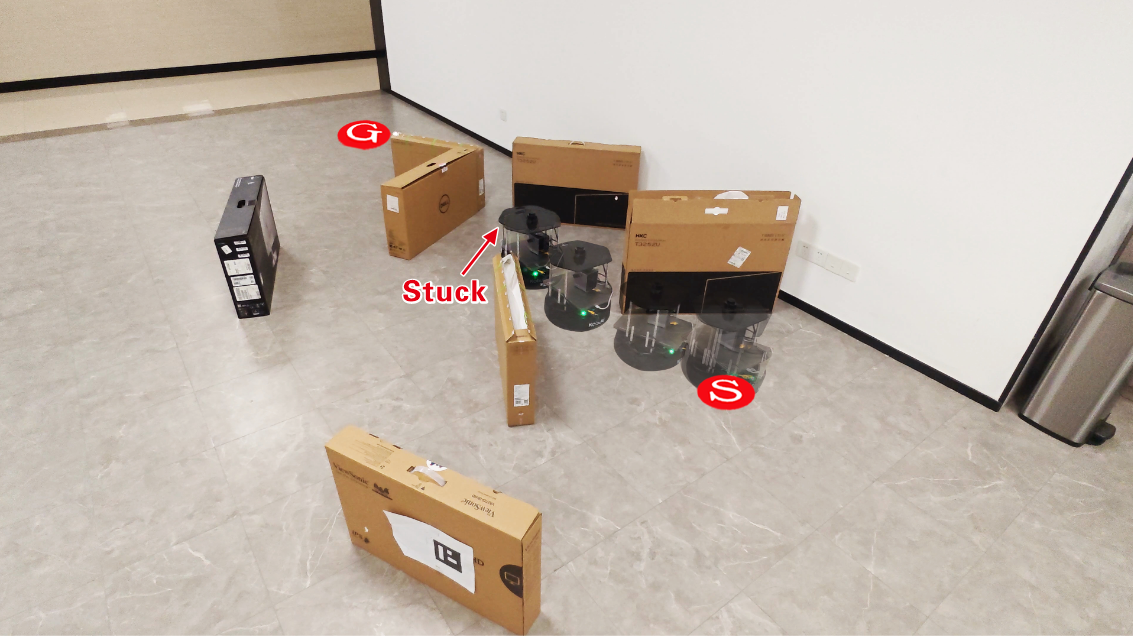}
        \caption{DRL-DCLP}  
        \label{static_drl}
    \end{subfigure}
    \begin{subfigure}[t]{0.60\columnwidth}
        \centering
        \includegraphics[width=\linewidth]{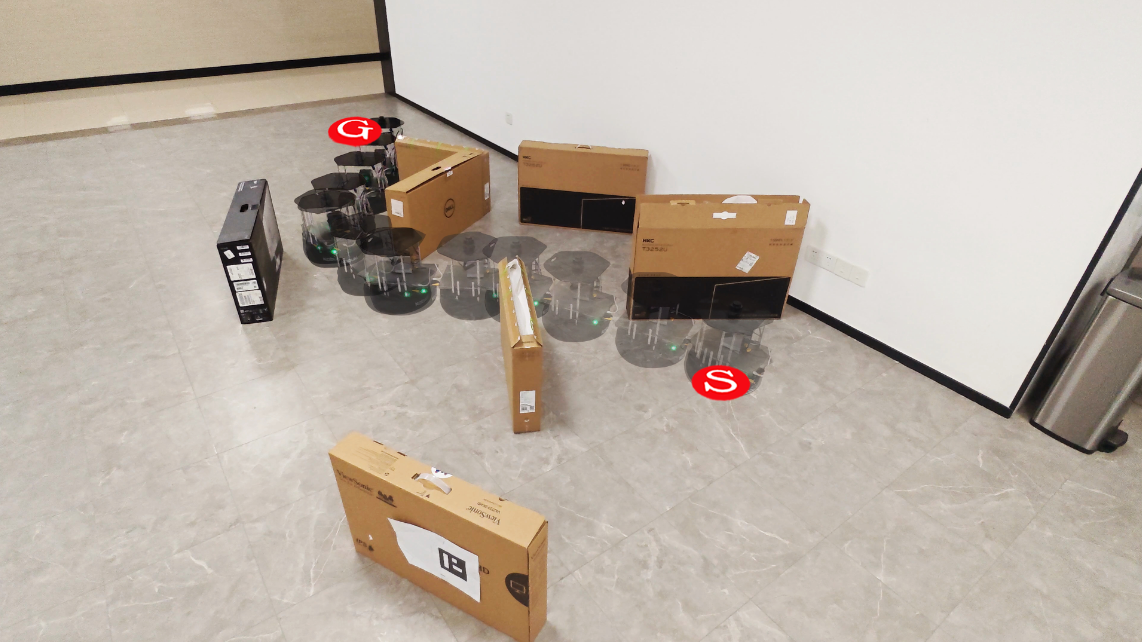}
        \caption{LME}
        \label{static_prl}
    \end{subfigure}
    \begin{subfigure}[t]{0.60\columnwidth}
        \centering
        \includegraphics[width=\linewidth]{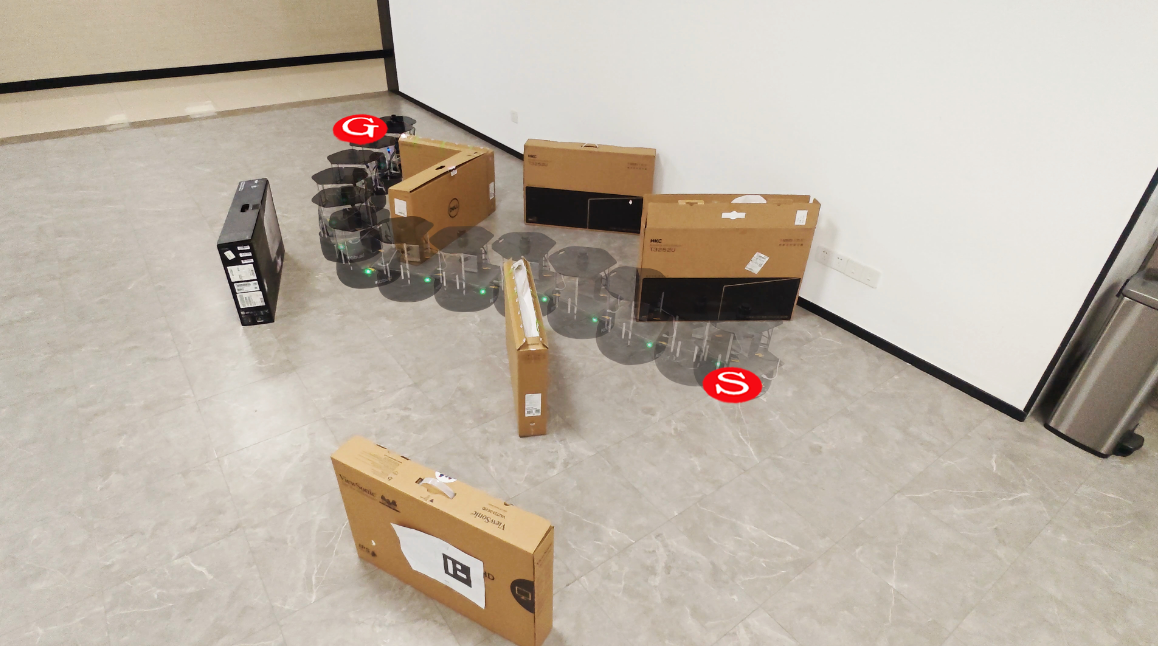}
        \caption{LME-DRL}
        \label{static_prl_drl}
    \end{subfigure}
    \caption{Trajectories of different methods in Real1. The starting point marked as ``S'' and the goal point marked as ``G''.}
    \label{real1}
	\vspace{-0.2cm}
\end{figure*}

\begin{figure*}[t]
    \centering
    \begin{subfigure}[t]{0.60\columnwidth}
        \centering
        \includegraphics[width=\linewidth]{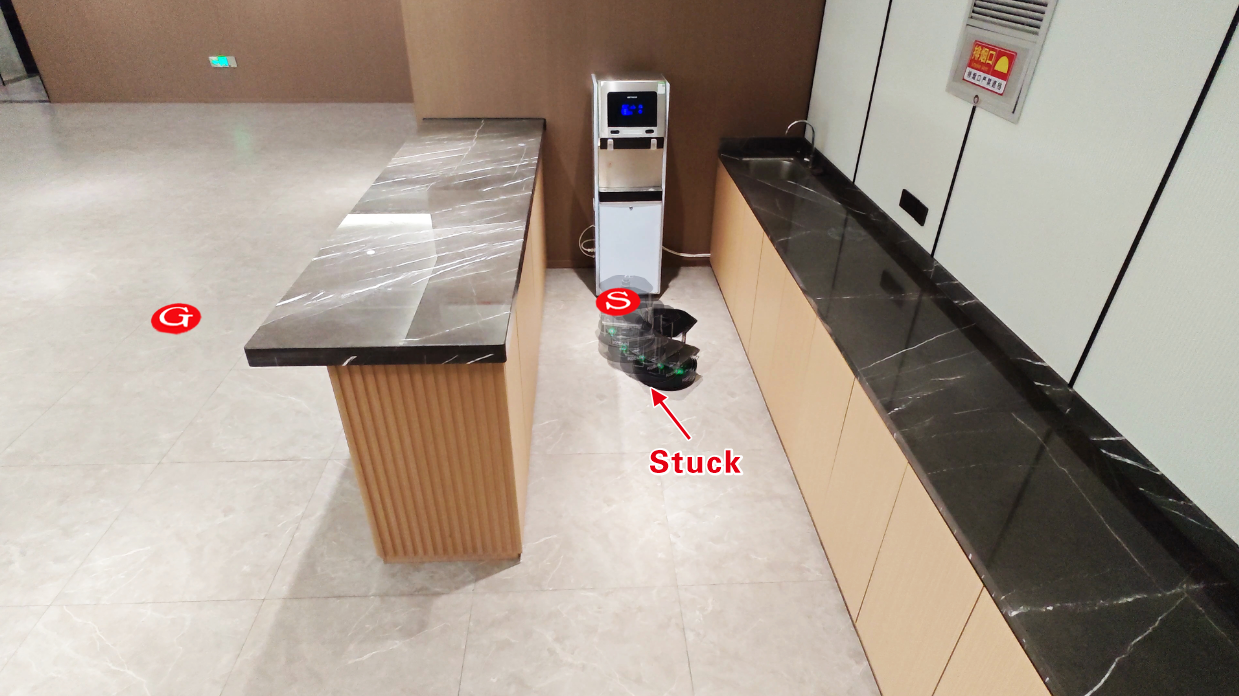}
        \caption{DRL-DCLP}
    \end{subfigure}
    \begin{subfigure}[t]{0.60\columnwidth}
        \centering
        \includegraphics[width=\linewidth]{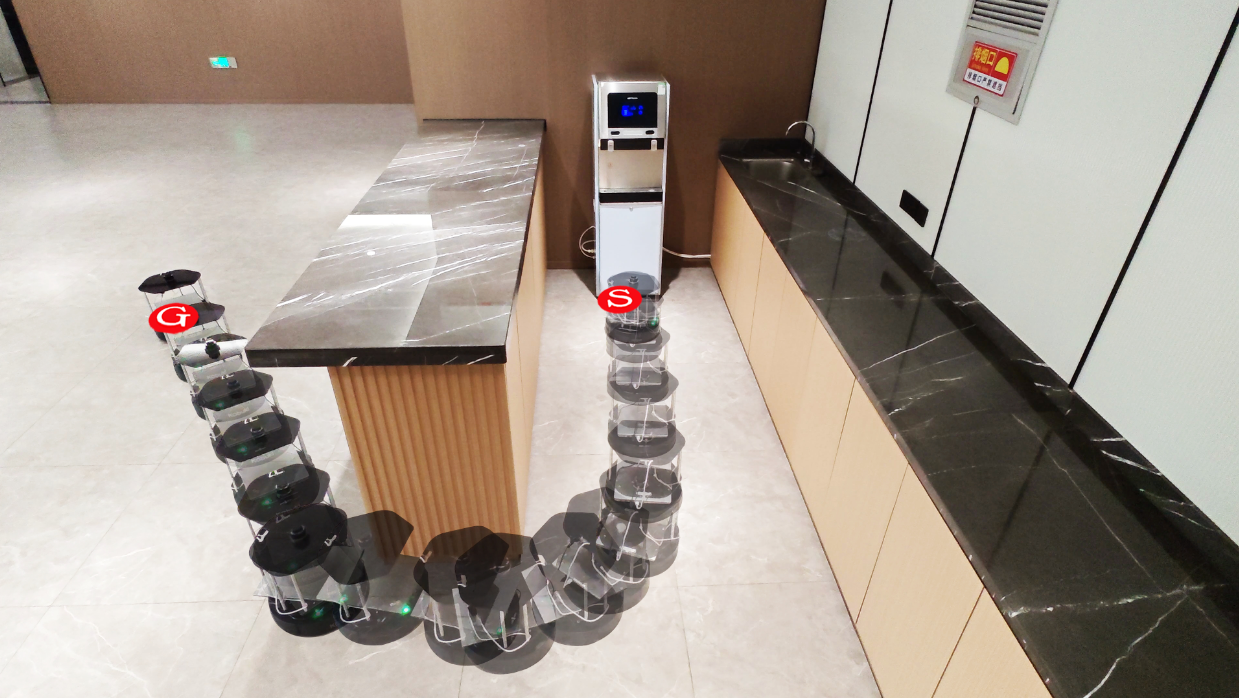}
        \caption{LME}
    \end{subfigure}
    \begin{subfigure}[t]{0.60\columnwidth}
        \centering
        \includegraphics[width=\linewidth]{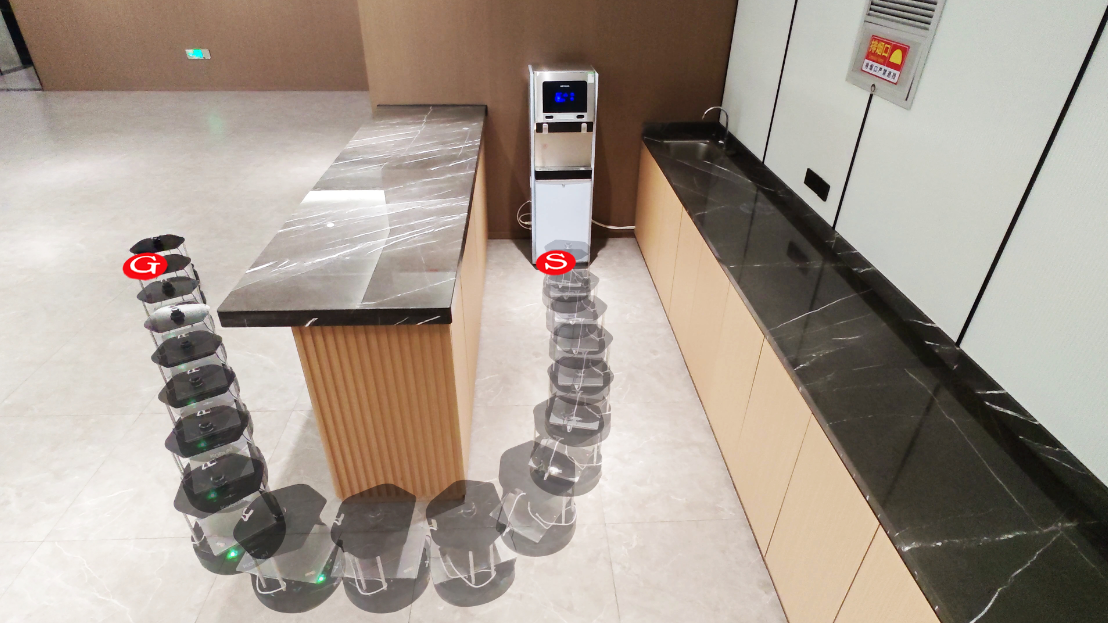}
        \caption{LME-DRL}
    \end{subfigure}
    \caption{Trajectories of different methods in Real2.}
    \label{real2}
\end{figure*}

\begin{figure*}[t]
    \centering
    \begin{subfigure}[t]{0.60\columnwidth}
        \centering
        \includegraphics[width=\linewidth]{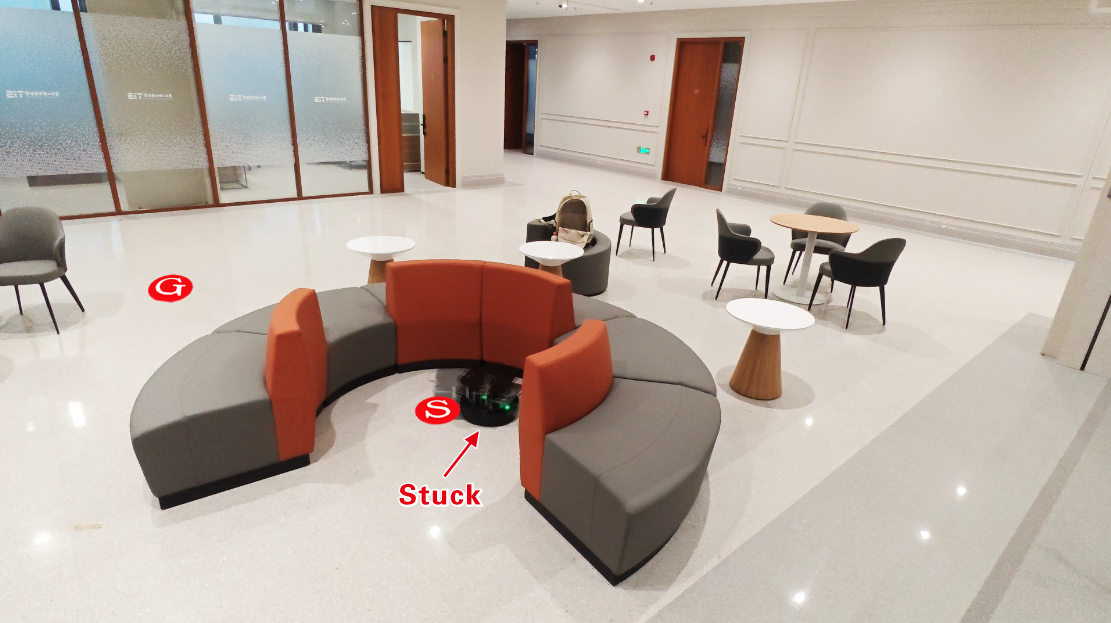}
        \caption{DRL-DCLP}  
    \end{subfigure}
    \begin{subfigure}[t]{0.60\columnwidth}
        \centering
        \includegraphics[width=\linewidth]{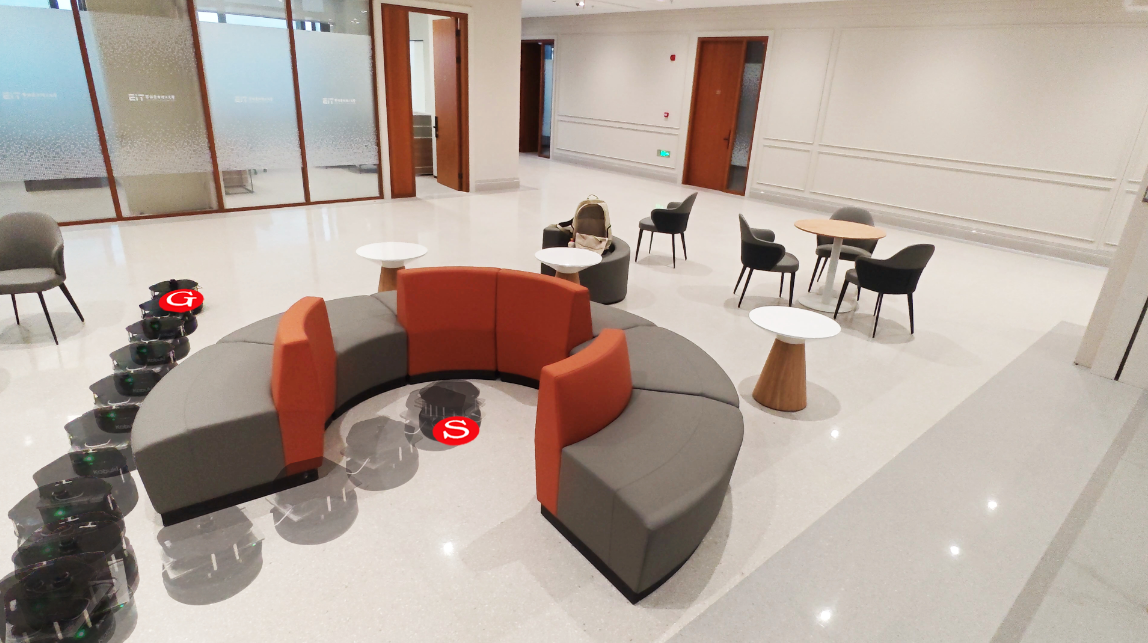}
        \caption{LME}
    \end{subfigure}
    \begin{subfigure}[t]{0.60\columnwidth}
        \centering
        \includegraphics[width=\linewidth]{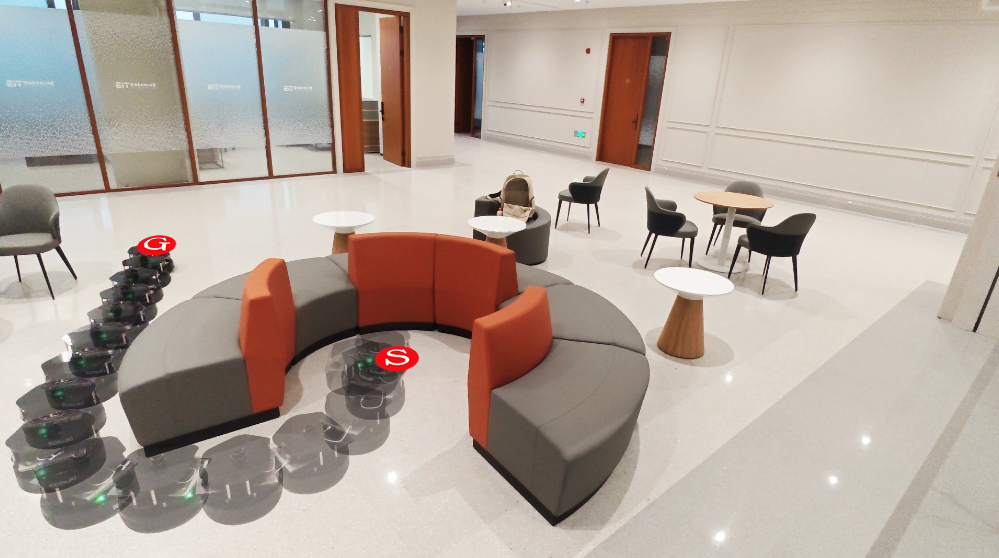}
        \caption{LME-DRL}
    \end{subfigure}
    \caption{Trajectories of different methods in Real3.}
    \label{real3}
\end{figure*}

\begin{figure*}[t]
    \centering
    \begin{subfigure}[t]{0.60\columnwidth}
        \centering
        \includegraphics[width=\linewidth]{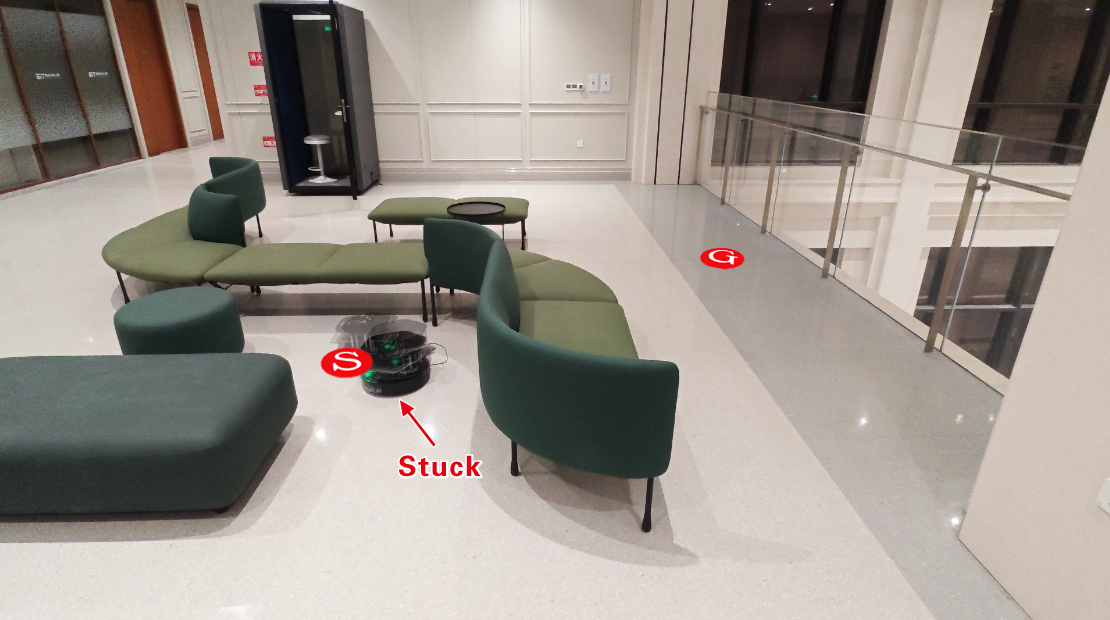}
        \caption{DRL-DCLP}  
    \end{subfigure}
    \begin{subfigure}[t]{0.60\columnwidth}
        \centering
        \includegraphics[width=\linewidth]{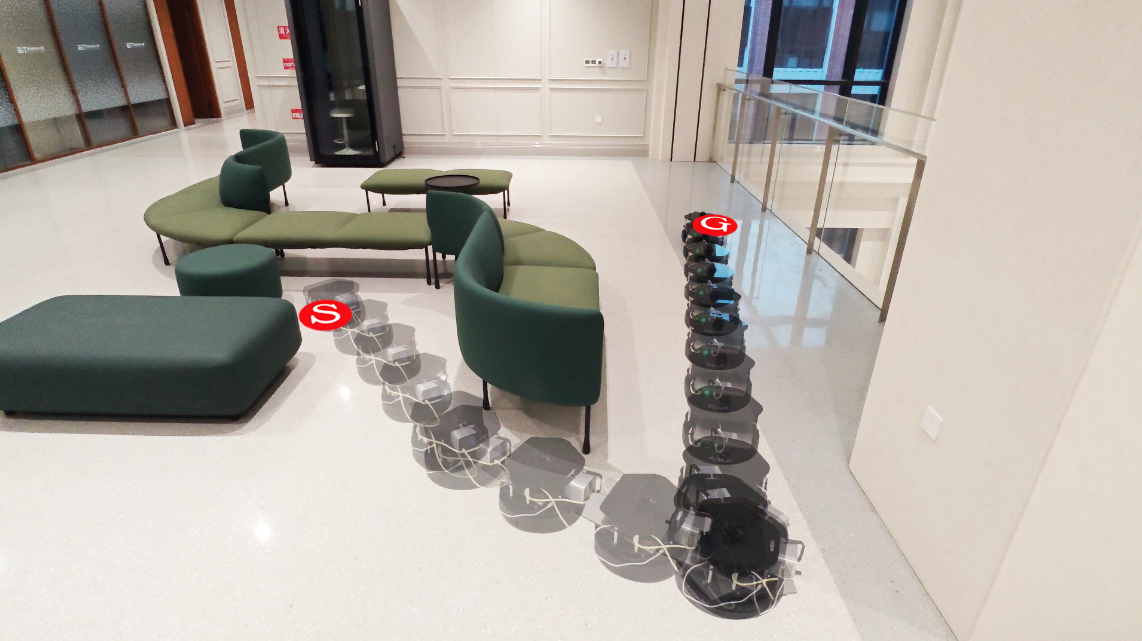}
        \caption{LME}
    \end{subfigure}
    \begin{subfigure}[t]{0.60\columnwidth}
        \centering
        \includegraphics[width=\linewidth]{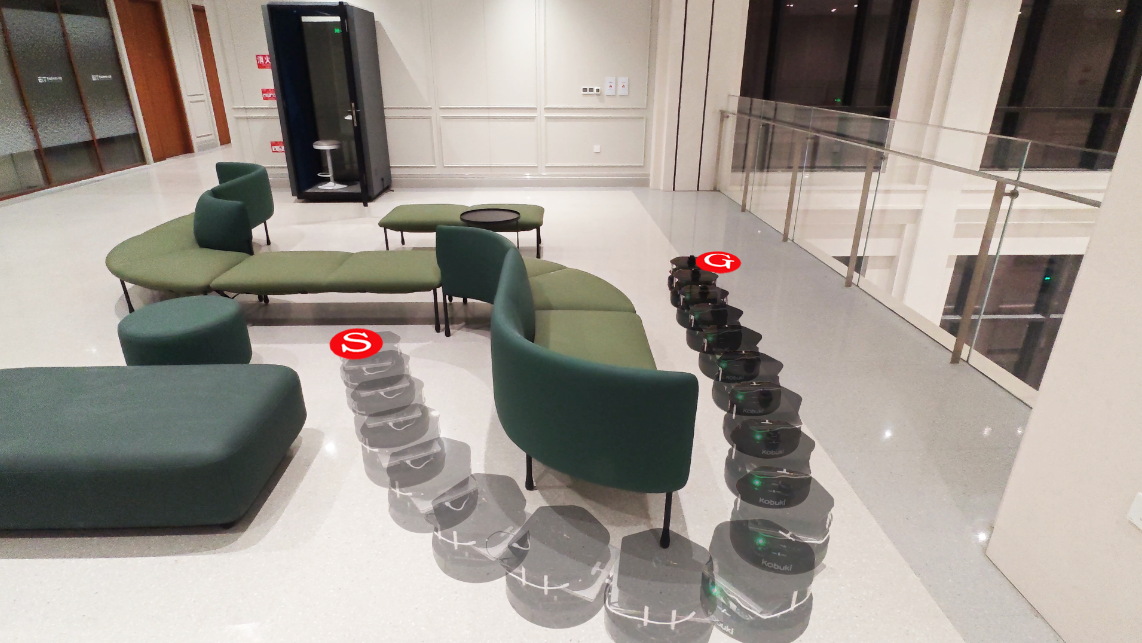}
        \caption{LME-DRL}
    \end{subfigure}
    \caption{Trajectories of different methods in Real4.}
    \label{real4}
	\vspace{-0.2cm}
\end{figure*}

\begin{figure*}[t]
    \centering
    \begin{subfigure}[t]{0.49\columnwidth}
        \centering
        \includegraphics[width=\linewidth]{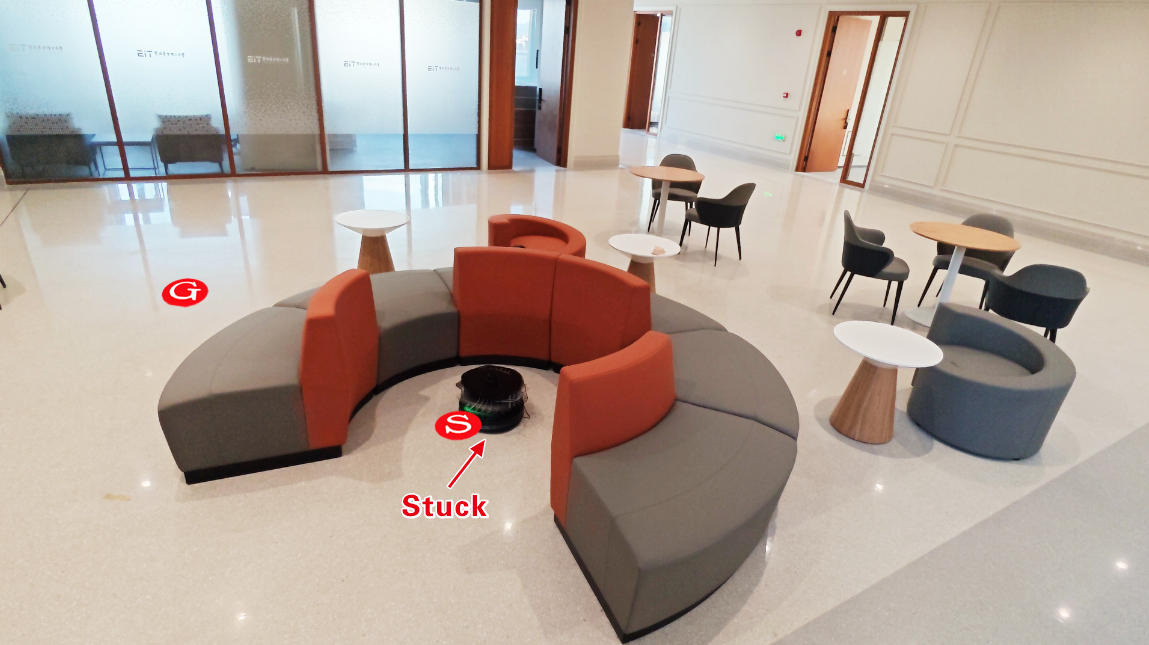}
        \caption{DWA}
    \end{subfigure}
    \begin{subfigure}[t]{0.49\columnwidth}
        \centering
        \includegraphics[width=\linewidth]{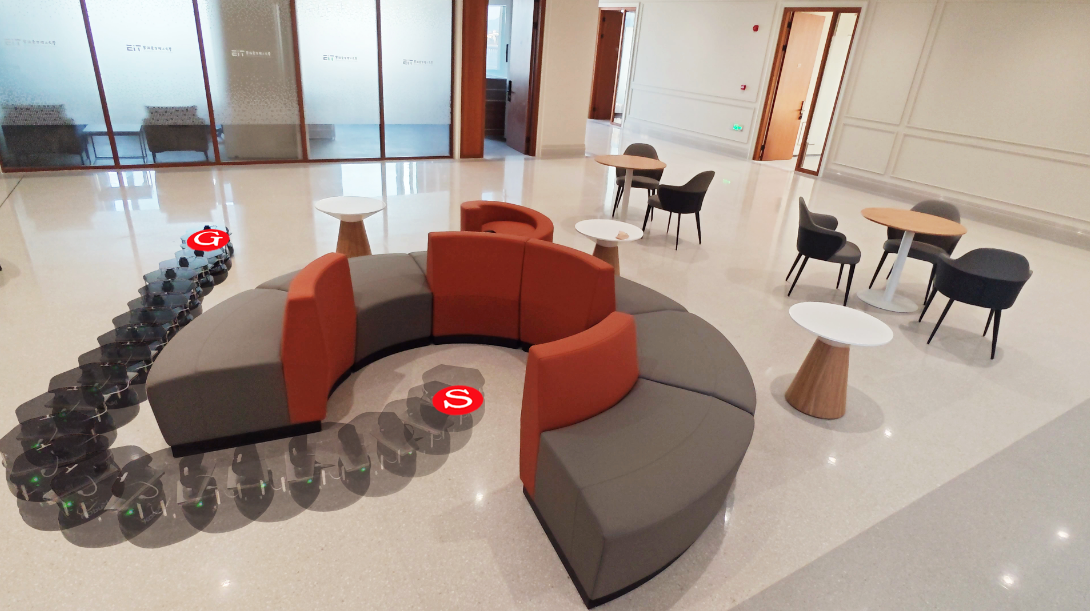}
        \caption{LME-DWA}
    \end{subfigure}
    \begin{subfigure}[t]{0.49\columnwidth}
        \centering
        \includegraphics[width=\linewidth]{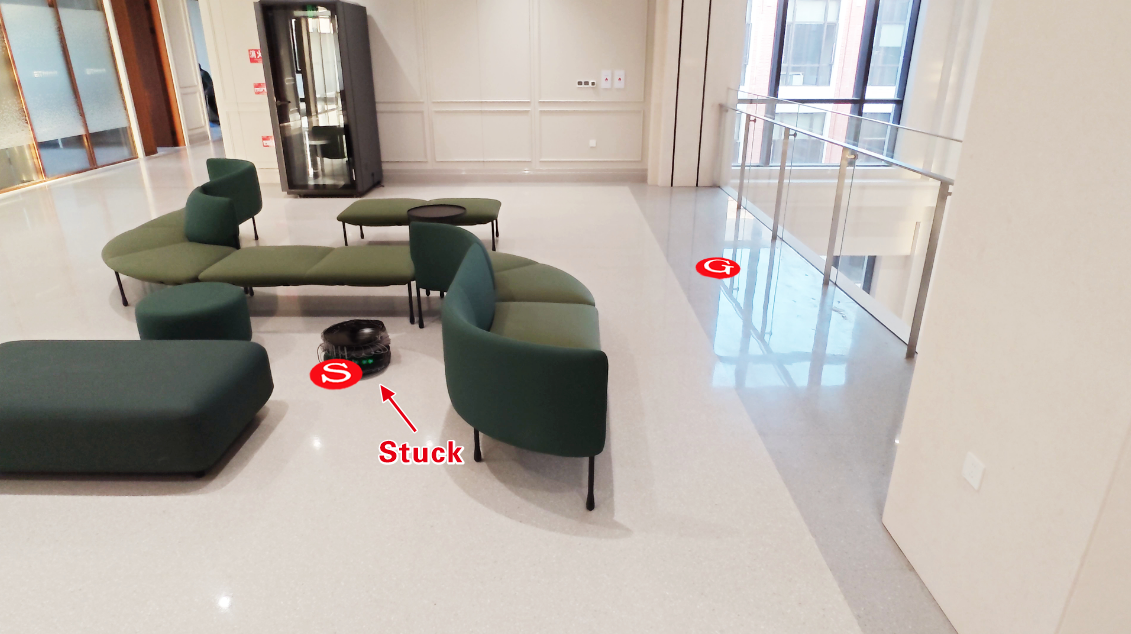}
        \caption{DWA}
    \end{subfigure}
    \begin{subfigure}[t]{0.49\columnwidth}
        \centering
        \includegraphics[width=\linewidth]{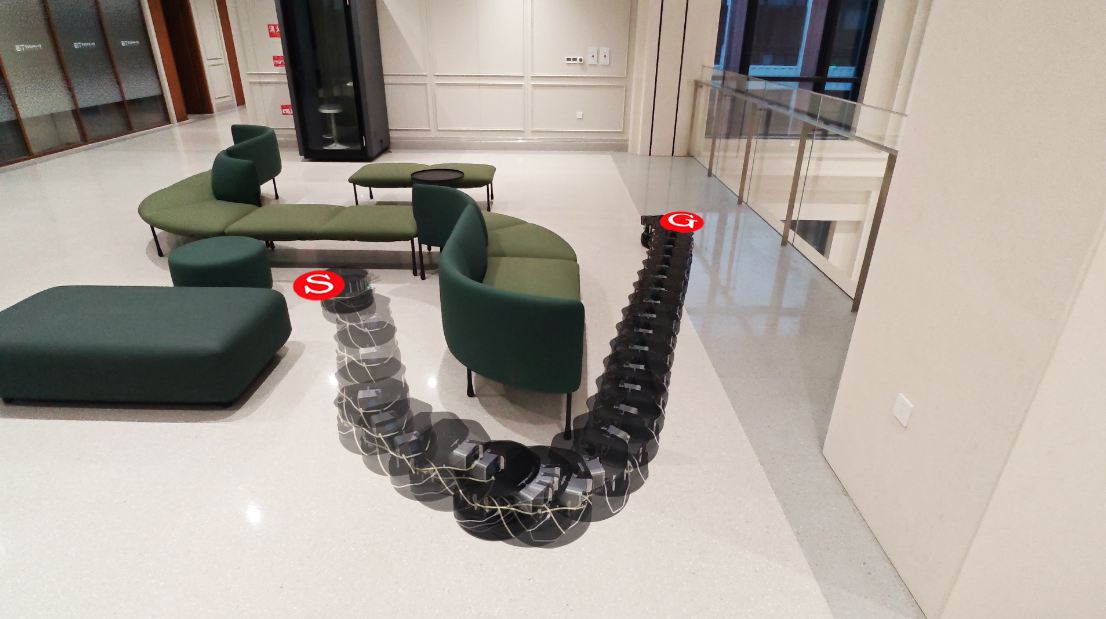}
        \caption{LME-DWA }
    \end{subfigure}
    \caption{Trajectories of DWA and LME-DWA.}
    \label{fig:dwa}
	\vspace{-0.2cm}
\end{figure*}

\section{Experiments in the Real World}

Given the competitive performance of our method across several simulation benchmarks, we deploy the LME policy in real-world scenarios to evaluate its effectiveness and generalization capability. The experiments include manually designed navigation scenarios as well as realistic environments containing local minima. To better reflect practical operating conditions, both static and dynamic obstacles are considered. We evaluate the method on two types of mobile robots: differential-drive wheeled robots and quadruped robots.

\subsection{Navigation Results for Wheeled Robots}

We directly deploy the LME policy trained in ROS Stage on a TurtleBot2 platform. A 360-degree Leishen N10P 2D LiDAR is mounted at the center of the robot, providing 270 uniformly distributed scan points obtained through linear interpolation. The robot is controlled by a laptop equipped with an Intel i5-8250U CPU without GPU acceleration. To obtain the relative position of the navigation target, we use ROS GMapping~\cite{Gmapping} to construct a map and ROS AMCL~\cite{AMCL} to localize the robot within the map. Importantly, the map is used only for target localization and is not provided as input to the policy.

\textbf{Static Obstacles.} We evaluate LME and the baseline methods in four real-world scenarios, as shown in Figs.~\ref{real1}--\ref{real4}. These scenarios, referred to as Real1--Real4, are designed to evaluate navigation robustness in environments with complex geometric structures. They include narrow passages, irregular obstacle configurations, concave corridor-like layouts, and semi-enclosed structures that can induce local minima. Such configurations are common in real-world environments and present substantial challenges to navigation methods relying primarily on local perception or short-horizon planning.

In Real1, DRL-DCLP successfully passes through the first narrow opening but becomes trapped at the second gap formed by two closely spaced obstacles. In contrast, LME identifies subgoals that guide the robot toward the boundaries of the obstacle configuration, enabling it to bypass the cluttered region. When coupled with the DRL local planner, LME-DRL similarly guides the robot out of the region where DRL-DCLP fails. These results demonstrate that the proposed method can effectively provide high-level guidance to local planners while maintaining collision-free navigation to the target.

In Real2--Real4, DRL-DCLP becomes trapped in the local-minimum regions, whereas both LME and LME-DRL successfully guide the robot toward the target. Notably, the trajectories generated by LME-DRL are smoother than those generated by LME alone, indicating that the combination of high-level subgoal guidance and local DRL control can improve trajectory smoothness. The trajectories obtained using DWA-based methods are presented in Fig.~\ref{fig:dwa}. Their overall performance is comparable to that of the DRL-based approaches.

\begin{figure*}[t]
    \centering
    \begin{subfigure}[t]{0.66\columnwidth}
        \centering
        \includegraphics[width=\linewidth]{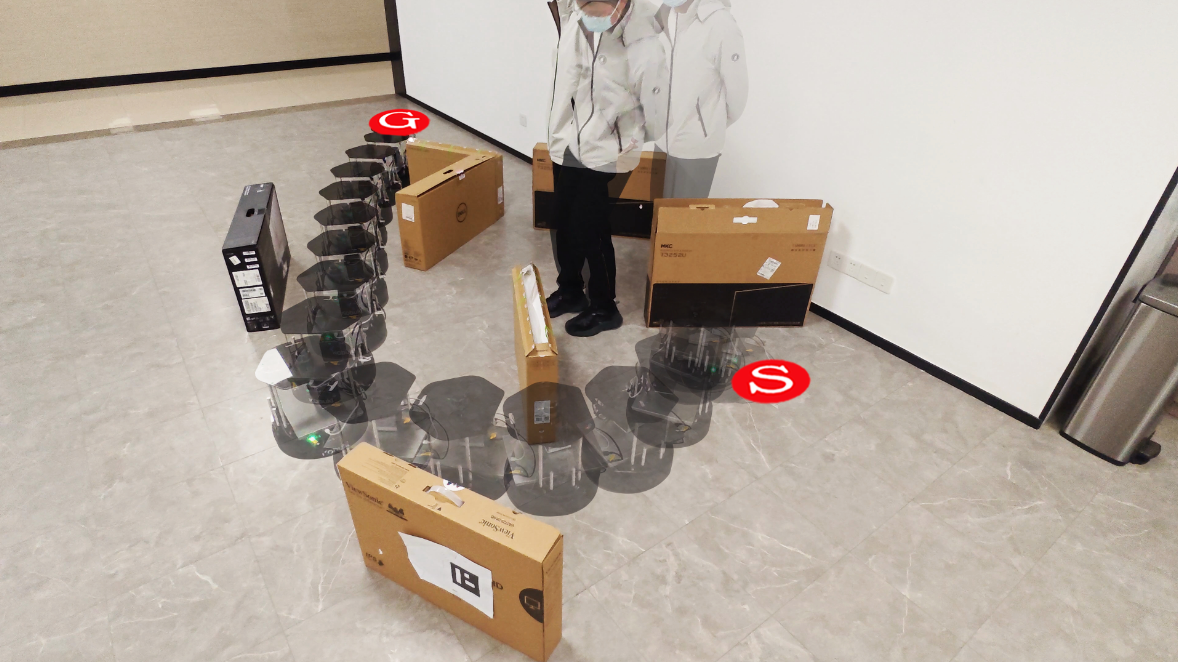}
        \caption{LME-DRL}
    \end{subfigure}
    \hspace{3mm}
    \begin{subfigure}[t]{0.66\columnwidth}
        \centering
        \includegraphics[width=\linewidth]{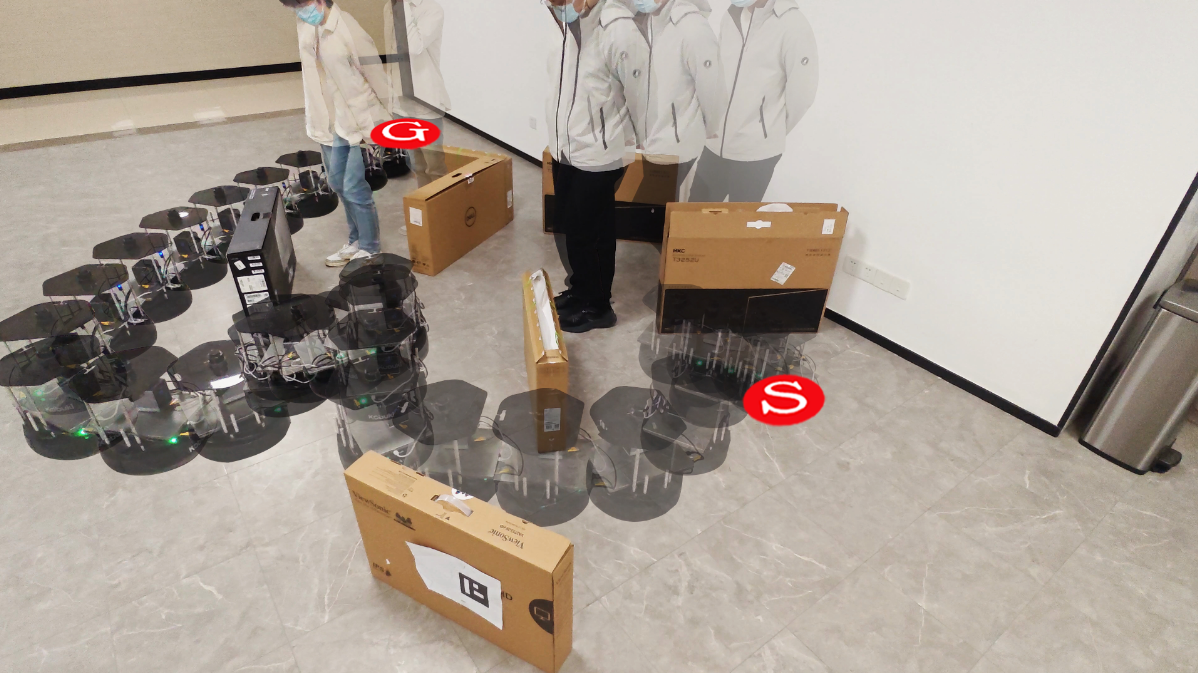}
        \caption{LME-DRL}
    \end{subfigure}
    \caption{Trajectories in dynamic obstacle scenarios.}
    \label{dynamic}
\end{figure*}

\begin{figure*}[t]
    \centering
    \begin{subfigure}[t]{0.68\columnwidth}
        \centering
        \includegraphics[width=\linewidth]{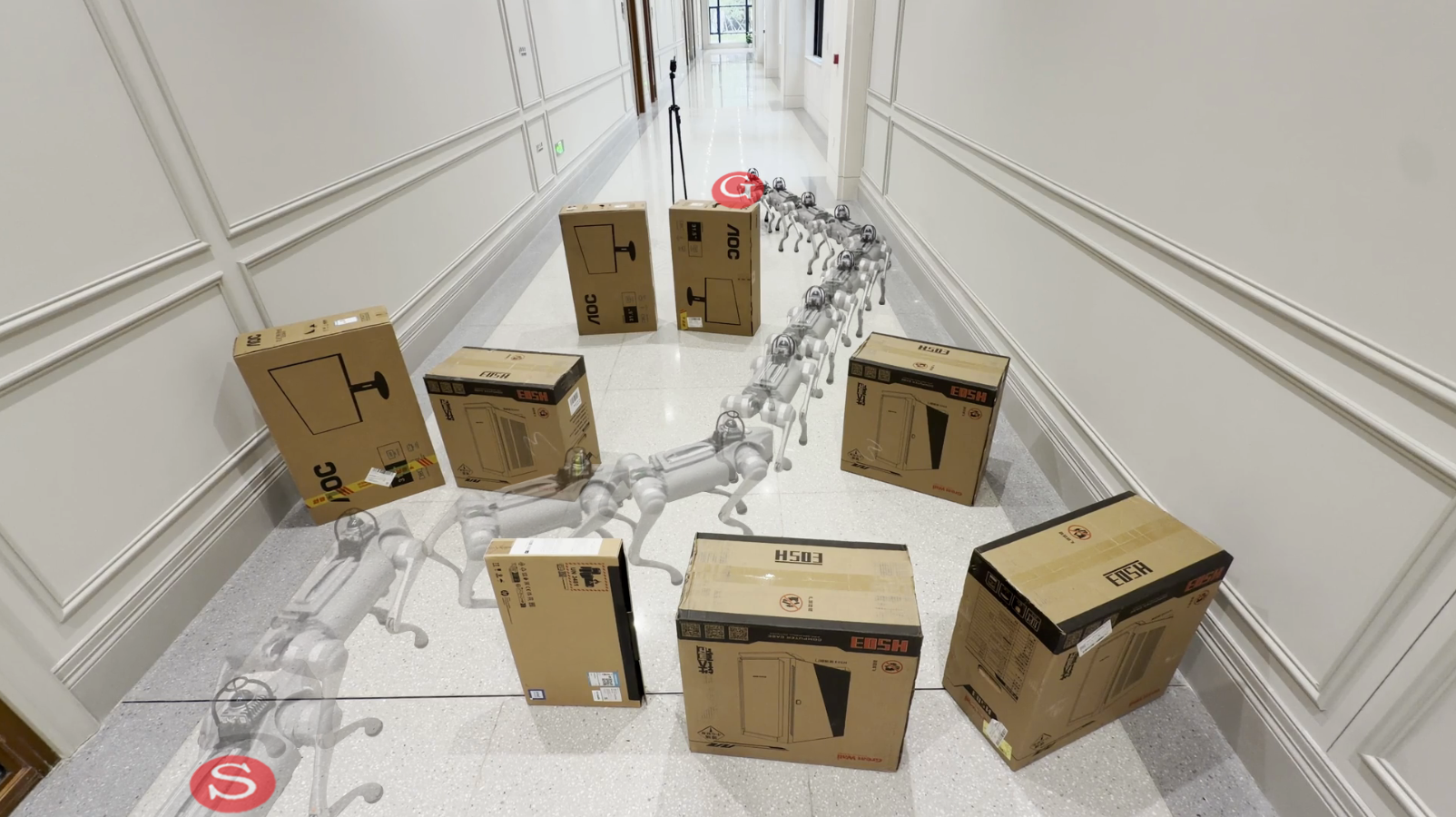}
        \caption{LME}
    \end{subfigure}
    \hspace{3mm}
    \begin{subfigure}[t]{0.68\columnwidth}
        \centering
        \includegraphics[width=\linewidth]{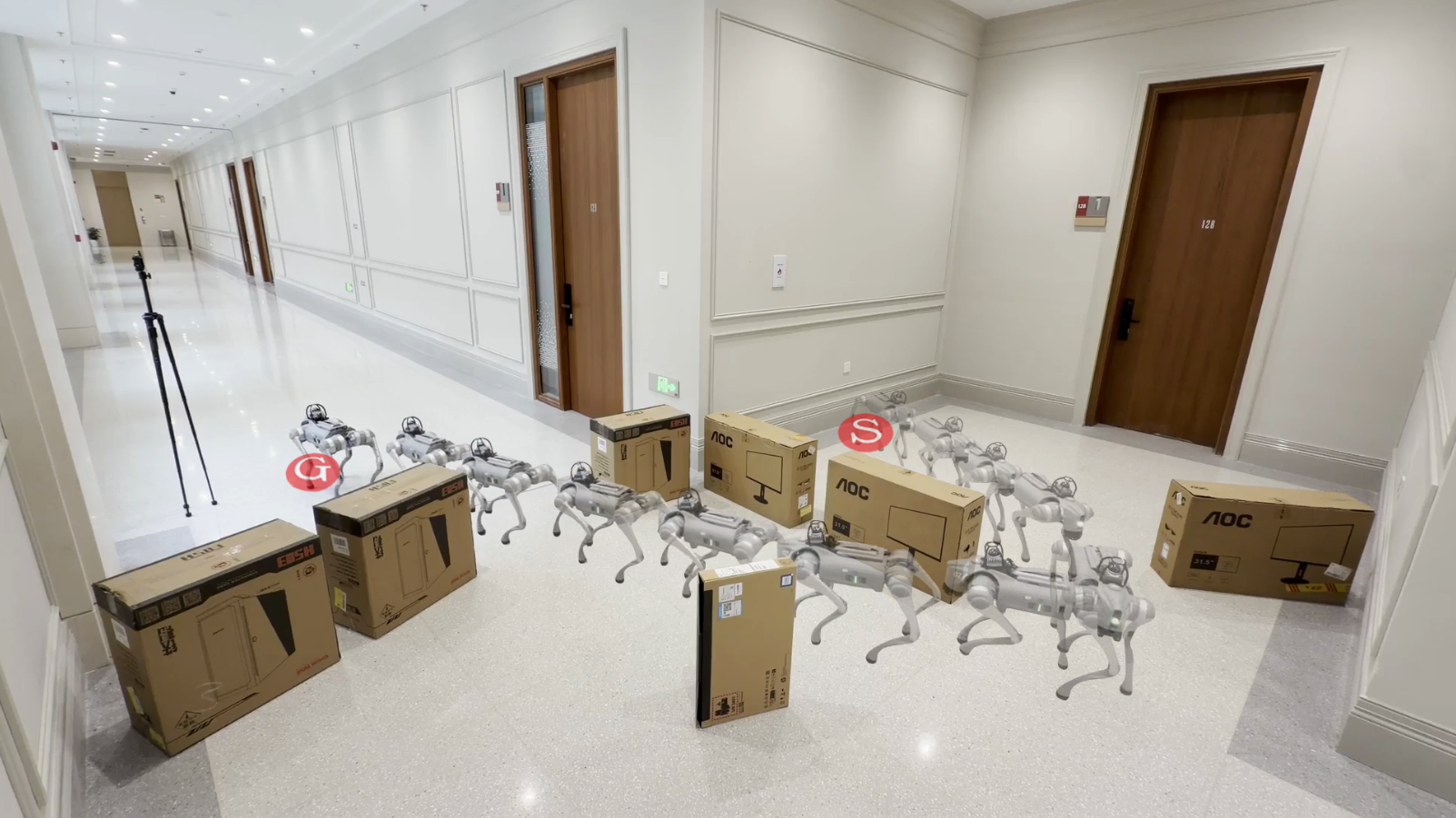}
        \caption{LME}
    \end{subfigure}
    \caption{Trajectories of Quadruped Robots.}
    \label{fig:dog}
	\vspace{-0.2cm}
\end{figure*}

\textbf{Dynamic Obstacle Scenarios.} Although dynamic obstacles are not considered during LME training, we additionally introduce dynamic obstacles into the Real1 scenario to evaluate the adaptability and generalization ability of our policy to unexpected obstacles. As illustrated in Fig.~\ref{dynamic}, when the robot approaches a narrow opening between obstacles, a pedestrian suddenly steps into the passage and blocks the original path. In such situations, the robot is expected to quickly replan an alternative path instead of getting stuck or colliding with the obstacle.

The resulting trajectories show that LME-DRL promptly changes its heading and successfully takes an alternative route. LME exhibits similar behavior. These results suggest that the subgoal generation module can rapidly generate new subgoals in response to changes in the environment, while the subgoal maintenance module switches to newly generated subgoals when the current subgoal becomes unsafe or unreachable. 

\subsection{Navigation Results for Quadruped Robots}  

We further deploy the policy on a Unitree Go2 quadruped robot to evaluate its applicability beyond differential-drive platforms. The robot is equipped with a Livox Mid-360 LiDAR for obstacle detection, and the resulting 3D point cloud is projected into a 2D laser scan to serve as input to the policy. An ultra-wideband (UWB) tag is placed at the target location to provide the relative target position. Onboard computation is performed using an NVIDIA Jetson Xavier NX. The policy operates at 8~Hz, with the maximum linear velocity limited to 0.66~m/s and the collision radius set to 0.3~m.

We design two challenging scenarios in which the robot must overcome local minima and avoid obstacles to reach the target. As shown in Fig.~\ref{fig:dog}, the proposed policy successfully navigates through both scenarios and reaches the target without collision. These experiments demonstrate that the proposed method can be transferred from simulation to a different real-world robotic platform and remains effective on quadruped robots with substantially different locomotion dynamics.

\section{Conclusion}

In this work, we proposed \textbf{LME} (Local-Minimum Escaper), a programmatic hierarchical framework for addressing local minima in unknown and partially observable environments. LME explicitly reasons about subgoals from local observations using interpretable heuristic criteria that jointly consider the geometric structure of surrounding obstacles and the safety of candidate locations. A local planner then generates low-level motion commands toward the selected subgoal. The resulting framework does not require a global map or additional training and remains independent of the underlying local planner.
Comprehensive experiments in both simulated and real-world environments demonstrate that LME can effectively generalize to challenging unseen scenarios, including highly constrained layouts, severe local-minimum configurations, and dynamic real-world environments. Furthermore, LME can seamlessly enhance different local planners by providing high-level subgoal guidance, enabling them to recover from local minima without additional training. The successful deployment on both differential-drive and quadrupedal robots further demonstrates the practical applicability and generality of the proposed framework. Overall, LME provides an interpretable approach for improving the robustness of local navigation systems in challenging environments.

\vspace{-0.1cm}

\bibliographystyle{IEEEtran}
\bibliography{ref.bib}



\end{document}